\documentclass[10pt,twocolumn,letterpaper]{article}

\usepackage{bm}
\usepackage{multirow}

\usepackage{cvpr}              

\usepackage[dvipsnames]{xcolor}
\newcommand{\red}[1]{{\color{red}#1}}

\definecolor{cvprblue}{rgb}{0.21,0.49,0.74}
\usepackage[pagebackref,breaklinks,colorlinks,citecolor=cvprblue]{hyperref}

\def\paperID{1038} 
\def\confName{CVPR}
\def\confYear{2024}

\title{
Learning Group Activity Features Through Person Attribute Prediction
}

\author{Chihiro Nakatani$^{1}$\quad Hiroaki Kawashima$^{2}$\quad Norimichi Ukita$^{1}$\\
$^{1}$ Toyota Technological Institute, Japan \quad $^{2}$ University of Hyogo, Japan}

\begin{document}
\maketitle
\begin{abstract}
  This paper proposes Group Activity Feature (GAF) learning in which features of multi-person activity are learned as a compact latent vector.
  Unlike prior work in which the manual annotation of group activities is required for supervised learning,
  our method learns the GAF through person attribute prediction without group activity annotations.
  By learning the whole network in an end-to-end manner so that the GAF is required for predicting the person attributes of people in a group, the GAF is trained as the features of multi-person activity.
  As a person attribute, we propose to use a person's action class and appearance features because the former is easy to annotate due to its simpleness, and the latter requires no manual annotation.
  In addition, we introduce a location-guided attribute prediction to disentangle the complex GAF for extracting the features of each target person properly.
  Various experimental results validate that our method outperforms SOTA methods quantitatively and qualitatively on two public datasets.
  Visualization of our GAF also demonstrates that our method learns the GAF representing fined-grained group activity classes.
  Code: \url{https://github.com/chihina/GAFL-CVPR2024}.
\end{abstract}    
\section{Introduction}
\label{sec:introduction}

A group activity is defined as what multiple people jointly engage in. 
Group activities are important targets in image and video understanding, such as team plays in sports~\cite{DBLP:conf/eccv/WuZBW22}, conversations in social scenes~\cite{DBLP:conf/iccvw/ChoiSS09}, and people flows in surveillance cameras~\cite{DBLP:conf/cvpr/SultaniCS18}.

\begin{figure}[t]
\begin{center}
\includegraphics[width=\columnwidth]{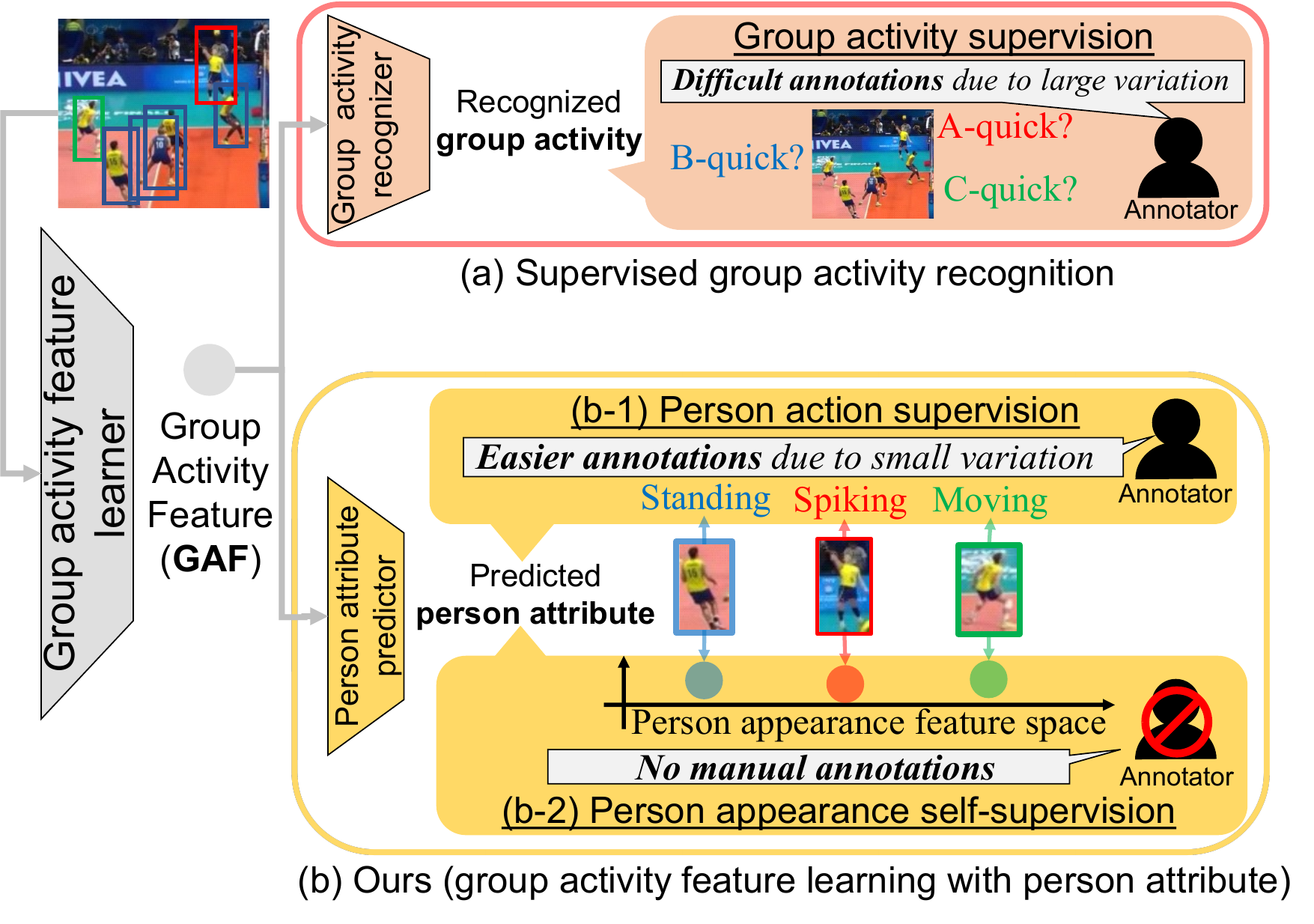}
\end{center}
\vspace*{-4mm}
\caption{
Difference between annotations for GAR and our group activity feature learning.
(a) Supervised GAR employs group activity annotations that are difficult due to various similar group activities.
(b-1) Our GAF learning employs person action annotations that are easy due to their simplicity. 
(b-2) We further propose annotation-free GAF learning with person appearance features. 
}
\label{fig:top}
\end{figure}

Group Activity Recognition ({\bf GAR}), in which a frame or video is classified into either of the predefined group activity classes, has been widely investigated
in recent years
~\cite{DBLP:conf/eccv/IbrahimM18,DBLP:conf/cvpr/WuWWGW19,DBLP:conf/cvpr/AzarANA19,DBLP:conf/eccv/PramonoCF20,DBLP:conf/eccv/EhsanpourASSRR20,DBLP:conf/cvpr/GavrilyukSJS20,DBLP:conf/iccv/Yuan0W21,DBLP:conf/iccv/LiCLYLHY21,DBLP:conf/cvpr/0002Z0Y0CQ22,DBLP:conf/eccv/TamuraVV22,DBLP:conf/eccv/ZhouKSGLZLKG22,DBLP:conf/cvpr/KimLCK22,DBLP:conf/iccvw/ChoiSS09,DBLP:conf/eccv/YanXTS020,DBLP:conf/mva/NakataniSU21,DBLP:conf/cvpr/XieGWC23,DBLP:journals/pami/YanXTST23}.
All of these GAR methods are based on supervised learning that requires the ground-truth group activities, as shown in Fig.~\ref{fig:top} (a).
In addition, it is known that person action recognition, which is also achieved based on a supervised learning manner, supports GAR~\cite{DBLP:conf/eccv/IbrahimM18,DBLP:conf/cvpr/WuWWGW19,DBLP:conf/cvpr/AzarANA19,DBLP:conf/eccv/PramonoCF20,DBLP:conf/eccv/EhsanpourASSRR20,DBLP:conf/cvpr/GavrilyukSJS20,DBLP:conf/iccv/LiCLYLHY21,DBLP:conf/cvpr/0002Z0Y0CQ22,DBLP:conf/eccv/TamuraVV22,DBLP:conf/eccv/ZhouKSGLZLKG22,DBLP:conf/iccvw/ChoiSS09,DBLP:journals/pami/YanXTST23,DBLP:conf/mva/NakataniSU21,DBLP:conf/cvpr/XieGWC23}.
%

Such supervised learning requires manually-annotated training data.
For GAR, group activity annotations are 
required.
In addition to labor-intensive and erroneous annotations,
a difficulty peculiar to GAR is the complexity of the group activities.
For example, while only four elemental group activity classes
are annotated in a widely-used team-sport dataset~\cite{DBLP:conf/cvpr/IbrahimMDVM16}, they may be insufficient for practical purposes such as tactical analysis (e.g., 200 or more plays are defined in American Football~\cite{bib:AFP2016}).
%
That is, more fine-grained activity classes are required for several applications of group activity analysis.
It is, however, difficult to correctly define and annotate such
complex, fine-grained activity classes, even manually, because of visually minor but highly contextual differences among those classes.

Such difficulty in manual annotations of group activities motivates us to learn a Group Activity Feature ({\bf GAF}) in which features of multi-person activity are learned as a compact latent vector without group activity annotations.
%
We note here that GAFs, which represent the complex features of multi-person activity, may have enough information to predict the attributes of each person in a group.

\begin{figure}[t]
\begin{center}
\includegraphics[width=\columnwidth]{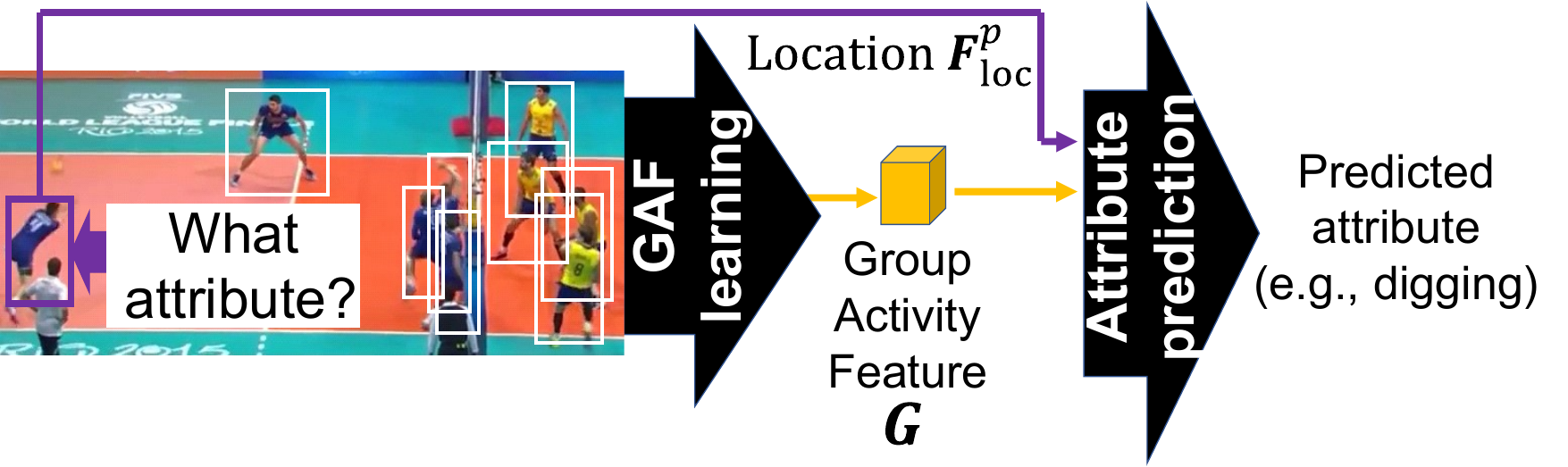}
\end{center}
\vspace*{-3mm}
\caption{Example of our group activity feature learning.
In this example, a group activity feature is learned to extract the scene context (i.e., spike group activity) through prediction of person attribute (e.g., digging).
See Fig.~\ref{fig:overview_network} for the detailed architecture.
}
\label{fig:overview_idea}
\end{figure}

With this regard, this paper proposes GAF learning using person attribute (e.g., action and appearance), which is easier to provide compared with the complex group activity annotation, as shown in Fig.~\ref{fig:top} (b).
%
%
With the person attribute related to the group activity, the GAF can narrow down the possible attribute of each person.
%
For example, Fig.~\ref{fig:overview_idea}, in which a spike is observed, shows that the possible person action enclosed by the purple rectangle (i.e., digging) can be narrowed down by the GAF representing the scene context (i.e., spike group activity) with the person's location.
%
%

As a person attribute for our GAF learning, person action can be used as shown in Fig.~\ref{fig:top} (b-1).
This is because person action is strongly related to the group activities and is essential to support GAR as mentioned above.
%
However, manual annotations are still required to use person action.
%
While the annotations of person action are easier than the one of group activities defined with complex people interaction, such action annotations are still labor-intensive.

To alleviate such difficulty in manual annotations, we also propose to learn GAF through the task of appearance feature prediction of each person in a group without manual annotations, as shown in Fig.~\ref{fig:top} (b-2). 
While the reduction of annotation cost have been studied widely, including
active learning~\cite{DBLP:conf/cvpr/RanaR23, DBLP:conf/eccv/HeilbronLJG18}, few-shot learning~\cite{DBLP:conf/eccv/ZhengCJ22,DBLP:conf/cvpr/WangZQTZG0S22}, and self-supervised learning~\cite{DBLP:conf/cvpr/LarssonMS17,DBLP:conf/iclr/GidarisSK18,DBLP:conf/cvpr/ZhangQWL19,DBLP:conf/iccv/DoerschGE15,DBLP:conf/eccv/NorooziF16,DBLP:conf/cvpr/PathakKDDE16,DBLP:conf/cvpr/DuanZCL22,DBLP:conf/cvpr/WangJBHLL19,DBLP:conf/cvpr/YaoLLZY20,DBLP:conf/eccv/WangJL20,DBLP:conf/cvpr/FernandoBGG17}, we follows the framework of self-supervised learning, where supervision signals are derived from the input data and any manual annotations are unnecessary.

Our novel contributions are summarized as follows:

\begin{itemize}
\item {\bf GAF learning through person attribute prediction:}
Unlike supervised GAR, we propose GAF learning through person attribute prediction without group activity annotations.
%
This paper proposes its two variants:
%
    \begin{itemize}
    \item {\bf GAF Learning with Person Action Classes (GAFL-PAC):}
    As a person attribute, we utilize person action classes, which are more easily annotated than group activities defined with complex inter-people interactions (Fig.~\ref{fig:top} (b-1)).
    In addition, for practical use of GAR and other group-related applications, the annotations of person actions are already given in general~\cite{DBLP:conf/eccv/IbrahimM18,DBLP:conf/iccvw/ChoiSS09}.
    \item {\bf GAF Learning with Person Appearance Features  (GAFL-PAF):}
    We also utilize person appearance features that can be obtained by a pre-trained model (e.g., VGG) without manual annotation (Fig.~\ref{fig:top} (b-2)).
    The appearance features are suitable as a person attribute due to the high relationship with multi-person activity.
    \end{itemize}

\item {\bf Fine granularity of our GAF:}
With the GAF learning through person attribute prediction, our method can learn fine-grained GAF that represents visual-subtle but important differences that are not represented in a manually-defined activity class, as shown in Fig.~\ref{fig:latent_sl_volley_examples}.

\item {\bf Location-Guided Person Attribute Prediction Using GAF:} 
While predicting the person attribute using the GAF is required in our method, extracting the features of each target person from the GAF is difficult.
This is because the GAF represents complex features of multi-person activity.
This feature extraction is achieved by location guidance in which each person's location feature is embedded into the GAF with positional encoding.

\end{itemize}

\section{Related Work}
\label{sec:related_work}

\subsection{Group Activity Recognition}
\label{subsection:gar}

\noindent{\bf Supervision by group activity labels.}
While we propose the GAF learning without group activity annotations (Fig.~\ref{fig:top} (b)), various GAR methods supervised by group activity annotations (i.e., Fig.~\ref{fig:top} (a)) have been proposed.
In~\cite{DBLP:conf/iccv/Yuan0W21,DBLP:conf/cvpr/KimLCK22,DBLP:journals/pami/YanXTST23}, only the labels of group activity are required for training. In~\cite{DBLP:conf/cvpr/KimLCK22}, a group activity is recognized from a whole image without any person features.
Kim~\textit{et al}~\cite{DBLP:conf/iccv/Yuan0W21} and Yan~\textit{et al}~\cite{DBLP:journals/pami/YanXTST23} employ the set of person features as input for a Graph Neural Network (GNN) as with~\cite{DBLP:conf/cvpr/WuWWGW19,DBLP:conf/eccv/EhsanpourASSRR20}.

\noindent{\bf Supervision by the labels of person action and group activity.}
Different from the aforementioned methods only with group activity supervision, the GAR network is jointly trained with the person action recognition network in~\cite{DBLP:conf/eccv/IbrahimM18,DBLP:conf/cvpr/WuWWGW19,DBLP:conf/cvpr/AzarANA19,DBLP:conf/eccv/PramonoCF20,DBLP:conf/eccv/EhsanpourASSRR20,DBLP:conf/cvpr/GavrilyukSJS20,DBLP:conf/iccv/LiCLYLHY21,DBLP:conf/cvpr/0002Z0Y0CQ22,DBLP:conf/eccv/TamuraVV22,DBLP:conf/eccv/ZhouKSGLZLKG22,DBLP:conf/iccvw/ChoiSS09,DBLP:conf/mva/NakataniSU21} for augmenting GAR.
In~\cite{DBLP:conf/cvpr/WuWWGW19,DBLP:conf/eccv/EhsanpourASSRR20}, GNN models the interactions between person features. In~\cite{DBLP:conf/cvpr/GavrilyukSJS20,DBLP:conf/iccv/LiCLYLHY21,DBLP:conf/cvpr/0002Z0Y0CQ22,DBLP:conf/eccv/TamuraVV22,DBLP:conf/eccv/ZhouKSGLZLKG22}, Transformer~\cite{DBLP:conf/nips/VaswaniSPUJGKP17} improves modeling the interactions between person features for GAR.

While all of the methods introduced in Sec.~\ref{subsection:gar} employ the annotation of group activities, our method learns GAF without group activity labels.
Following the success of interaction modeling in~\cite{DBLP:conf/cvpr/GavrilyukSJS20,DBLP:conf/iccv/LiCLYLHY21,DBLP:conf/cvpr/0002Z0Y0CQ22,DBLP:conf/eccv/TamuraVV22,DBLP:conf/eccv/ZhouKSGLZLKG22}, we also learn the GAF with Transformer.

\subsection{Self-supervised Representation Learning}
\label{subsection:self}

\noindent{\bf Image-based pretext tasks.}
Most pretext tasks for self-supervised image representation utilize the transformation of original images.
In Gidaris~\textit{et al}.~\cite{DBLP:conf/iclr/GidarisSK18}, Zhang~\textit{et al}.~\cite{DBLP:conf/cvpr/ZhangQWL19}, and Larsson~\textit{et al}.~\cite{DBLP:conf/cvpr/LarssonMS17}, each original image is rotated, affinely- and projectively-transformed, and grayscaled, respectively.
The image representation model is trained so that the model undoes each image.
While these methods~\cite{DBLP:conf/iclr/GidarisSK18,DBLP:conf/cvpr/ZhangQWL19,DBLP:conf/cvpr/LarssonMS17} employ a whole image, patches extracted from the image are used in~\cite{DBLP:conf/iccv/DoerschGE15,DBLP:conf/eccv/NorooziF16,DBLP:conf/cvpr/PathakKDDE16}. 
Doersch~\textit{et al}.~\cite{DBLP:conf/iccv/DoerschGE15} predict the spatial configuration (i.e., relative positions) of randomly-sampled patches. 
In Noroozi~\textit{et al}.~\cite{DBLP:conf/eccv/NorooziF16}, jigsaw puzzles in which all patches of an image are shuffled are solved.
Pathak~\textit{et al}.~\cite{DBLP:conf/cvpr/PathakKDDE16} inpaint partially-erased images.

\noindent{\bf Video-based pretext tasks.}
While the image-based methods can be applied to a video, video-specific pretext tasks are also proposed in~\cite{DBLP:conf/cvpr/DuanZCL22,DBLP:conf/cvpr/WangJBHLL19,DBLP:conf/cvpr/YaoLLZY20,DBLP:conf/eccv/WangJL20,DBLP:conf/cvpr/FernandoBGG17}.
In Yao~\textit{et al}.~\cite{DBLP:conf/cvpr/YaoLLZY20} and Wang~\textit{et al}.~\cite{DBLP:conf/eccv/WangJL20}, the paces of video clips
are arbitrarily changed (e.g., normal, half, and double speeds), and video representation is learned by solving a pace prediction problem.
In Fernando~\textit{et al}.~\cite{DBLP:conf/cvpr/FernandoBGG17}, sequential video clips are randomly reordered, and video representation is learned so that these reordered and original clips can be classified.
%
Wang~\textit{et al}.~\cite{DBLP:conf/cvpr/WangJBHLL19} employ appearance cues as well as motion statistics for spatial-temporal representation learning. 

While all of the representation learning methods introduced above extract low-level image features for various downstream tasks such as general image/video classification and prediction, our method focuses on features useful for group-related tasks  (e.g., group scene retrieval and clustering).
As with our method, Ibrahim~\textit{et al}.~\cite{DBLP:conf/eccv/IbrahimM18} aims to extract multi-person scene features through GNN in an unsupervised manner,
while the relation of each person to a group activity is not fully captured.
%
Our method incorporates such intimate relations between individuals and group activity (e.g., digging in a spike scene) using the task of location-guided person attribute prediction from a GAF.

\begin{figure*}[t]
\begin{center}
\includegraphics[width=\textwidth]{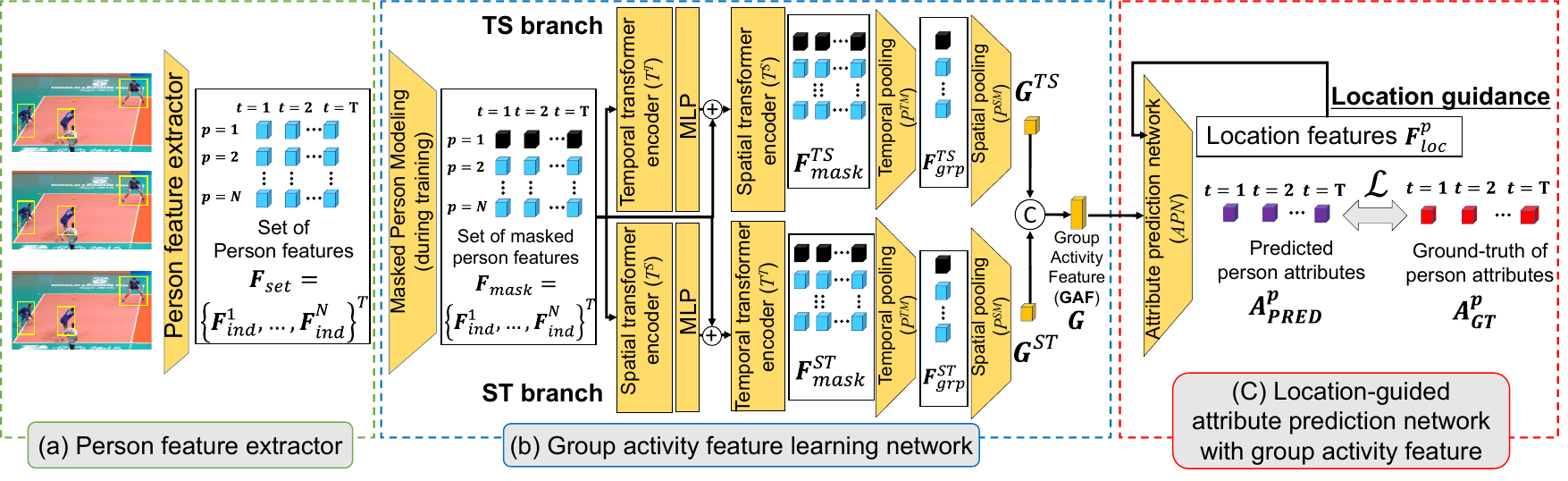}\\
\end{center}
\vspace{-2mm}
\caption{Overview of our GAF learning network. (a) Person feature extractor. The person feature is composed of appearance and location features. (b) GAF learning network. The GAF is learned from extracted people features. (c) Location-guided attribute prediction network with the GAF. The attribute of each person is predicted from the location feature of the person and the GAF extracted in  (b).}
\label{fig:overview_network}
\end{figure*}
\section{Proposed Method}
\label{sec:poposed_method}


Our GAF learning network consists of three stages, (a), (b), and (c) (Fig.~\ref{fig:overview_network}).
In stage (a), the features of each person are extracted (Sec.~\ref{subsec:person_feature_extractor}).
In stage (b), the features of several people are masked (i.e., removed) during training for GAF enhancement (Sec.~\ref{subsec:masked_person_modeling}).
Then, the masked person features are fed into the transformer-based GAF learning network  (Sec.~\ref{subsec:group_feature_extractor}).
In stage (c), the attribute of each person is predicted from the GAF with location guidance (Sec.~\ref{subsec:attribute_prediction}).

\subsection{Person Feature Extractor}
\label{subsec:person_feature_extractor}

\paragraph{Overview.}
As shown in Fig.~\ref{fig:overview_network} (a), the set of person features $\bm{F}_{set} \in\mathbb{R}^{T \times N \times C}$ are extracted from images. 
$\bm{F}_{set}$ is composed of the features of $N$ people obtained from $T$ frames in a video. 
$C$ is the dimension of person $p$'s feature vector, $\bm{F}^{p}_{ind} \in\mathbb{R}^{C}$, indicated by a blue cuboid in Fig.~\ref{fig:overview_network} (a).
$\bm{F}^{p}_{ind}$ consists of appearance features (denoted by $\bm{F}^{p}_{app} \in\mathbb{R}^{C}$) and location features (denoted by $\bm{F}^{p}_{loc} \in\mathbb{R}^{C}$)
in our implementation 
because their effectiveness is validated for GAR~\cite{DBLP:conf/eccv/PramonoCF20,DBLP:conf/cvpr/GavrilyukSJS20,DBLP:conf/iccv/LiCLYLHY21,DBLP:conf/cvpr/0002Z0Y0CQ22,DBLP:conf/eccv/ZhouKSGLZLKG22}, while other features (e.g., body keypoints) can also be used.

\paragraph{Detail.}
$\bm{F}^{p}_{app}$ is extracted by the following three steps, as with~\cite{DBLP:conf/iccv/Yuan0W21}.
First, a feature map is extracted from the whole image by VGG~\cite{DBLP:journals/corr/SimonyanZ14a}.
Then, RoIAlign~\cite{DBLP:conf/iccv/HeGDG17} is applied with the bounding box of each person to obtain the feature map for each person.
Finally, the feature map of each person is embedded into $C$-dimensional appearance features with a linear transformation.
From the $\bm{F}^{p}_{app}$ of $N$ people between $T$ frames, we construct $\bm{F}_{app} \in\mathbb{R}^{T \times N \times C}$.
In addition to $\bm{F}_{app}$, $\bm{F}^{p}_{loc}$ is also essential to understand the spatial structure of a group, as $\bm{F}^{p}_{loc}$ is used for GAR~\cite{DBLP:conf/eccv/PramonoCF20,DBLP:conf/cvpr/0002Z0Y0CQ22,DBLP:conf/cvpr/WuWWGW19,DBLP:conf/cvpr/GavrilyukSJS20,DBLP:conf/eccv/ZhouKSGLZLKG22}. 
As with~\cite{DBLP:conf/cvpr/0002Z0Y0CQ22}, the center point of each person bounding box, $(x, y)$, is embedded into $\bm{F}^{p}_{loc}$ by a spatial positional encoding
so that $\bm{F}_{loc} \in\mathbb{R}^{T \times N \times C}$ whose dimension is equal to $\bm{F}_{app}$.
%
Finally, $\bm{F}_{loc}$ is elementwise added with $\bm{F}_{app}$ to obtain the set of person features, $\bm{F}_{set}$.
Here, we denote person $p$'s features in $T$ frames (i.e., a slice of $\bm{F}_{set}$) as $\bm{F}_{ind}^{p} \in \mathbb{R}^{T \times C} \, (p \in \{1, \cdots, N\})$ and, similarly, person $p$'s location features in $T$ frames (i.e., a slice of $\bm{F}_{loc}$) as $\bm{F}_{loc}^{p} \in \mathbb{R}^{T \times C} \, (p \in \{1, \cdots, N\})$.

\subsection{Masked Person Modeling (MPM)}
\label{subsec:masked_person_modeling}

\paragraph{Overview.}
In the set of person features $\bm{F}_{set}$, the features of randomly sampled people are masked (e.g., $\bm{F}_{ind}^{1}$ indicated by a black cuboid in Fig.~\ref{fig:overview_network} (b)) during training.
%
%
%
By extracting such features with our Masked Person Modeling (MPM), our network is expected to learn the features of the interaction between unmasked people for predicting the attribute of the masked person.

\paragraph{Detail.}
The binary mask for $p$-th person (denoted by $\bm{M}^{p} \in\mathbb{R}^{T \times C}$) is initialized by filling 1 in all values. 
Then, $N_{mask}$ people to be masked are randomly sampled from all $N$ people in a scene.
All values in $\bm{M}^{p}$ for the randomly sampled people are updated with 0.
%
%

From the all $\bm{M}^{p}$ of $N$ people, we construct $\bm{M}=[M^{1},\cdots,M^{N}] \in\mathbb{R}^{T \times N \times C}$.
%
%
%
Then, $\bm{M}$ is elementwise multiplied by $\bm{F}_{set}$ to obtain the masked person features $\bm{F}_{mask}$ for our GAF training.
Note that all values in $\bm{M}$ are set to be 0 during inference to preserve features of all people in a GAF.

\subsection{Group Activity Feature Learning Network}
\label{subsec:group_feature_extractor}

\paragraph{Overview.}
The GAF $\bm{G}$ is extracted from the set of masked person features $\bm{F}_{mask}$ obtained in Sec.~\ref{subsec:masked_person_modeling}.
For this GAF learning, we employ Transformer to model spatial- and temporal-interactions between people in accordance with~\cite{DBLP:conf/iccv/LiCLYLHY21,DBLP:conf/cvpr/0002Z0Y0CQ22}.
Our transformer-based GAF learning network consists of two branches (i.e., TS and ST branches) similar to~\cite{DBLP:conf/cvpr/0002Z0Y0CQ22}.
As shown in Fig.~\ref{fig:overview_network}, the Temporal Transformer encoder ($T^{T}$) and the Spatial Transformer encoder ($T^{S}$) are used in both branches, but they are placed in reverse order in the two branches.
The combination of these two branches can model the spatial-temporal interactions between people, as proven in GAR~\cite{DBLP:conf/cvpr/0002Z0Y0CQ22,DBLP:conf/iccv/LiCLYLHY21}.
%
While the transformer-based network is used in our method, the network can be replaced with any SOTA network without difficulties because the stage (i.e., Fig.~\ref{fig:overview_network} (b)) is modularized.

\paragraph{Detail.}
$\bm{F}_{mask}$ is independently fed into the TS and ST branches to acquire $\bm{F}_{mask}^{TS}$ and $\bm{F}_{mask}^{ST}$, respectively:
\begin{align}
  \bm{F}_{mask}^{TS} &= T^{S}(\bm{F}_{mask}+MLP(T^{T}(\bm{F}_{mask})))\\
  \bm{F}_{mask}^{ST} &= T^{T}(\bm{F}_{mask}+MLP(T^{S}(\bm{F}_{mask})))
  \label{eq:person_feat_extraction}
\end{align}
where $MLP$ denotes the multi-layer perceptron.
$\bm{G}^{TS} \in\mathbb{R}^{C}$ and $\bm{G}^{ST} \in\mathbb{R}^{C}$ are obtained from $\bm{F}_{mask}^{TS}$ and $\bm{F}_{mask}^{ST}$, respectively, by the Temporal Max Pooling (denoted by $P^{TM}$) and the Spatial Max Pooling (denoted by $P^{SM}$), as with previous methods~\cite{DBLP:conf/cvpr/GavrilyukSJS20,DBLP:conf/cvpr/WuWWGW19}:
\begin{align}
  \bm{G}^{TS} &= P^{SM}(P^{TM}(\bm{F}_{mask}^{TS}))\\
  \bm{G}^{ST} &= P^{SM}(P^{TM}(\bm{F}_{mask}^{ST}))
  \label{eq:group_feat_extraction}
\end{align}
Finally, $\bm{G}^{TS}$ and $\bm{G}^{ST}$ are concatenated to obtain the final GAF $\bm{G} \in\mathbb{R}^{2C}$ as follows:
\begin{equation}
 \bm{G} = \bm{G}^{TS} \oplus \bm{G}^{ST}
 \label{eq:group_feat_concat}
\end{equation}

During training, $\bm{G}$ is fed into the attribute prediction network (denoted by $APN$), as shown in Fig.~\ref{fig:overview_network} (c).
%
By backpropagating a loss used in this person attribute prediction not only inside $APN$ but also across the whole network shown in Fig.~\ref{fig:overview_network}, $\bm{G}$ can be trained as the GAF.
The details of person attribute prediction and the loss used in this prediction are described in Sec.~\ref{subsec:attribute_prediction} and Sec.\ref{subsec:loss}, respectively.

After predicting $\bm{G}$ in inference, $\bm{G}$ can be used in various ways such as a pretrained model for downstream supervised tasks and unsupervised learning tasks (e.g., retrieval and clustering), as mentioned in Sec.~\ref{sec:introduction}.
The effectiveness of $\bm{G}$ in these tasks is validated in Sec.~\ref{sec:experiments}.

\subsection{Attribute Prediction with GAF}
\label{subsec:attribute_prediction}

\paragraph{Overview.}
Using $\bm{G}$ obtained in Sec.~\ref{subsec:group_feature_extractor}, the attribute of each person is predicted by the $APN$ (Fig.~\ref{fig:overview_network} (c)).
To predict the attribute of $p$-th person, we use their location features (i.e., $\bm{F}^{p}_{loc}$) as guidance for attribute prediction from $\bm{G}$.

\paragraph{Detail.}
$\bm{G}$ is fed into the attribute prediction network with the location of each person as follows:
\begin{align}
  \bm{A}_{PRED}^{p} &= APN(\bm{G},\bm{F}_{loc}^{p})
  \label{eq:att_rec_with_pos}
\end{align}
where $\bm{A}_{PRED}^{p}\in\mathbb{R}^{T\times R}$ denotes the predicted attribute of each person obtained from $\bm{G}$ with their location. 
The dimension of $R$ changes depending on the type of the predicted person attribute as follows:

\noindent {\bf (i) Person action:}
$R$ is the number of action classes and  the $A_{PRED}^{p}$ represents predicted action probabilities.

\noindent {\bf (ii) Person appearance features:}
$R$ is the dimension of the appearance features (i.e., $C$) extracted in Sec.~\ref{subsec:person_feature_extractor}.

\subsection{Loss Function}
\label{subsec:loss}

The whole network is trained with a loss function (denoted by $\mathcal{L}$) for attribute prediction as a pretext task as follows:

\noindent{\bf (i) Person action:}
When the person attribute is the action, $\mathcal{L}$ is the cross-entropy loss: $\mathcal{L} = \mathcal{L}_{CE}(\bm{A}_{PRED}^{p}, \bm{A}_{GT}^{p})$, where $\bm{A}_{GT}^{p}$ denotes the ground-truth one-hot vector for action.

\noindent{\bf (ii) Person appearance features:}
For appearance features, $\mathcal{L}$ is the mean squared loss function: $\mathcal{L} = \mathcal{L}_{MSE}(\bm{A}_{PRED}^{p}, \bm{A}_{GT}^{p})$, where $\bm{A}_{GT}^{p}$ denotes the extracted appearance features (i.e., $\bm{F}_{app}$) shown in Sec.~\ref{subsec:person_feature_extractor}.
\section{Experiments}
\label{sec:experiments}

\subsection{Datasets}
\label{subsection:datasets}

%
The Volleyball dataset, which contains highly correlated players, is mainly used to validate the effectiveness of our GAF learning.
The Collective Activity dataset is also used to validate the generality of our method in Sec.~\ref{subsection:comp_exp}, while people in this dataset are not highly related to each other compared with the Volleyball dataset.

\noindent{\bf VolleyBall Dataset (VBD)}~\cite{DBLP:conf/cvpr/IbrahimMDVM16} consists of 4,830 sequences extracted from 55 games. 
Each sequence is annotated with one of the predefined eight group activity classes, i.e., Left-spike, Right-spike, Left-set, Right-set, Left-pass, Right-pass, Left-winpoint, and Right-winpoint.
While each sequence has 41 frames, 
its center 20 frames
have annotations with the full-body bounding boxes of all players and their action classes, i.e., Waiting, Setting, Digging, Falling, Spiking, Jumping, Moving, Blocking, and Standing.

\noindent{\bf Collective Activity Dataset (CAD)}~\cite{DBLP:conf/iccvw/ChoiSS09} contains 44 videos.
In each video, every ten frames are annotated with person action classes, i.e., NA, Crossing, Waiting, Queuing, Walking, and Talking, and their bounding boxes. 
The group activity class is determined by the largest number of person actions in 
each frame,
while the NA class is not included as a group activity class.
We follow the previous methods~\cite{DBLP:conf/iccv/Yuan0W21,DBLP:conf/cvpr/WangNY17} to merge the Crossing and Walking into Moving.

\begin{table*}[t]
    \centering
    \caption{
    Quantitative comparison of retrieval on the VolleyBall Dataset (VBD).
    The results obtained in two experimental settings (i.g., GAFL-PAC and GAFL-PAF) are separated by double lines.
    The best result in each column is colored in \textcolor{red}{red}.
    Results obtained by the concatenation of output features (i.e., $\bm{F}^{TS}_{grp}$ and $\bm{F}^{ST}_{grp}$) and $\bm{G}$ are denoted as ``Ours-ind'' and ``Ours-grp'', respectively.
    }
    \vspace{-2mm}
    {\small
    \begin{tabular}{c|l||r|r|r|r|r|r|r|r|r|r|r} \hline
    & Retrieval type & \multicolumn{4}{c|}{Action set (IoU~\cite{DBLP:conf/eccv/IbrahimM18})} & \multicolumn{4}{c|}{Action set (AF-IDF)} & \multicolumn{3}{c}{Group activity}\\ \hline
    & Method & Hit@1 & Hit@2 & Hit@3 & mAP & Hit@1 & Hit@2 & Hit@3 & mAP & Hit@1 & Hit@2 & Hit@3 \\\hline \hline
    
    \multirow{5}{*}{\shortstack{GAFL\\-\\PAC}} & HiGCIN~\cite{DBLP:journals/pami/YanXTST23} & 74.3 & 84.9 & 89.5 & 55.7 & 59.8 & 73.6 & 80.3 & 30.5 & 50.0 & 66.3 & 74.5 \\ \cline{2-13}
    & DIN~\cite{DBLP:conf/iccv/Yuan0W21} & 79.7 & 90.1 & 93.4 & 60.2 & 74.5 & 85.2 & 88.3 & 39.3 & 57.0 & 73.1 & 81.1 \\ \cline{2-13}
    & Dual-AI~\cite{DBLP:conf/cvpr/0002Z0Y0CQ22} & 67.6 & 84.7 & 91.6 & 56.9 & 72.6 & 83.7 & 88.6 & 53.0 & 64.4 & 76.5 & 82.0 \\ \cline{2-13}
    & Ours-ind & 82.7 & 91.6 & 95.0 & 59.1 & 79.0 & 86.8 & 89.8 & 45.6 & 82.7 & 88.8 & 91.3 \\ \cline{2-13}
    & Ours-grp & \red{83.0} & \red{92.7} & \red{95.5} & \red{64.2} & \red{80.1} & \red{88.4} & \red{91.5} & \red{59.9} & \red{84.8} & \red{89.6} & \red{91.8} \\ \hline\hline
    
    \multirow{5}{*}{\shortstack{GAFL\\-\\PAF}} & B1-Compact128~\cite{DBLP:conf/eccv/IbrahimM18} & 57.9 & 75.7 & 84.3 & 45.8 & 41.3 & 60.8 & 71.4 & 29.3 & 30.3 & 48.0 & 59.9 \\ \cline{2-13}
    & B2-VGG19~\cite{DBLP:conf/eccv/IbrahimM18} & 63.8 & 80.6 & 86.8 & 46.8 & 46.7 & 65.8 & 75.7 & 29.4 & 35.4 & 53.6 & 65.0 \\ \cline{2-13}
    & HRN~\cite{DBLP:conf/eccv/IbrahimM18} & 60.9 & 78.6 & 86.0 & \red{46.9} & 40.8 & 60.9 & 72.9 & 28.7 & 31.2 & 47.0 & 57.6 \\ \cline{2-13}
    & Ours-ind & 64.2 & 80.8 & 88.3 & 45.0 & 50.4 & 69.3 & 77.6 & 30.1 & 55.0 & 72.3 & 79.2 \\ \cline{2-13}
    & Ours-grp & \red{64.8} & \red{82.7} & \red{90.3} & 46.4 & \red{52.3} & \red{71.4} & \red{81.0} & \red{31.4} & \red{61.1} & \red{75.1} & \red{82.4} \\ \hline
    \end{tabular}
    }
    \label{table:comparison_volley}
    \centering
    \vspace{1mm}
    \caption{Quantitative comparison of retrieval on the Collective Activity Dataset (CAD).
    }
    \vspace{-3mm}
    {\small
    \begin{tabular}{c|l||r|r|r|r|r|r|r|r|r|r|r} \hline
    & Retrieval type & \multicolumn{4}{c|}{Action set (IoU~\cite{DBLP:conf/eccv/IbrahimM18})} & \multicolumn{4}{c|}{Action set (AF-IDF)} & \multicolumn{3}{c}{Group activity}\\ \hline
    & Method & Hit@1 & Hit@2 & Hit@3 & mAP & Hit@1 & Hit@2 & Hit@3 & mAP & Hit@1 & Hit@2 & Hit@3 \\\hline\hline

    \multirow{5}{*}{\shortstack{GAFL\\-\\PAC}} & HiGCIN~\cite{DBLP:journals/pami/YanXTST23} & 80.8 & 85.4 & 89.7 & 57.9 & 81.0 & 85.2 & 89.3 & 61.6 & 86.1 & 88.8 & 91.9 \\ \cline{2-13}
    & DIN~\cite{DBLP:conf/iccv/Yuan0W21} & 71.4 & 74.1 & 74.9 & 51.5 & 90.1 & 92.7 & 94.0 & 52.8 & 90.8 & 92.5 & 93.2 \\ \cline{2-13}
    & Dual-AI~\cite{DBLP:conf/cvpr/0002Z0Y0CQ22} & 61.0 & 72.5 & 76.7 & 61.5 & 85.5 & 86.9 & 88.1 & 82.7 & 82.1 & 84.1 & 84.7 \\ \cline{2-13}
    & Ours-ind & 76.2 & 82.6 & 89.4 & \red{78.9} & 94.8 & 95.6 & 95.9 & 82.2 & \red{94.9} & 95.4 & 95.7 \\ \cline{2-13}
    & Ours-grp & \red{81.8} & \red{90.7} & \red{93.5} & 69.9 & \red{96.1} & \red{96.5} & \red{96.6} & \red{93.9} & \red{94.9} & \red{95.6} & \red{96.3} \\ \hline\hline
    
    \multirow{5}{*}{\shortstack{GAFL\\-\\PAF}}  & B1-Compact128~\cite{DBLP:conf/eccv/IbrahimM18} & 48.8 & 60.3 & 68.2 & 38.0 & 81.8 & 88.2 & 89.7 & 52.6 & 82.4 & 88.4 & 90.1 \\ \cline{2-13}
    & B2-VGG19~\cite{DBLP:conf/eccv/IbrahimM18} & 53.6 & 61.6 & 66.1 & 35.3 & 71.1 & 80.3 & 83.8 & 46.7 & 72.2 & 80.8 & 84.2 \\ \cline{2-13}
    & HRN~\cite{DBLP:conf/eccv/IbrahimM18} & 37.1 & 50.1 & 58.6 & 22.2 & 53.2 & 64.8 & 72.5 & 34.2 & 54.0 & 64.8 & 72.4 \\ \cline{2-13}
    & Ours-ind & \red{67.6} & \red{81.3} & \red{85.9} & \red{53.3} & \red{83.7} & \red{88.9} & \red{90.2} & 57.5 & \red{88.5} & \red{91.2} & \red{91.9} \\  \cline{2-13}
    & Ours-grp & 52.7 & 70.3 & 74.1 & 46.4 & 74.0 & 80.5 & 82.6 & \red{60.1} & 79.2 & 81.0 & 82.0 \\ \hline
    \end{tabular}
    }
    \label{table:comparison_cad}
\end{table*}

\subsection{Evaluation Protocols}
\label{subsection:evaluation}

\subsubsection{Evaluation Tasks}
\label{subsubsection:scene_matching}

The quality of the estimated GAF is verified with the following two types of retrieval tasks.

\noindent{\bf Action set retrieval.}
We employ the same action set retrieval task as in~\cite{DBLP:conf/eccv/IbrahimM18}.
Action IoU in~\cite{DBLP:conf/eccv/IbrahimM18} evaluates the similarity of action structures between query and retrieved images based on the overlap of action distributions.
If the IoU exceeds a predefined threshold (e.g., 0.5), the query and retrieved images are regarded as matched.
%
However, in this action IoU, all action classes are counted equally without weights, although the action class distribution is imbalanced (e.g., the percentage of ``standing'' people, who are less informative for tactics, is over 68\% in VBD).

To resolve this class imbalanced problem, we propose Action Frequency-Inverse Scene Frequency (AF-ISF) inspired by Term Frequency-Inverse Document
Frequency (TF-IDF~\cite{DBLP:journals/ipm/Aizawa03}) that evaluates the importance of a word in a document.
In AF-ISF, each scene is represented by a feature vector in which each value is computed from frequency statics of action classes as follows:
\begin{align}
  \bm{FV}^{i} = [FV^{i}_{1},FV^{i}_{m},\cdots,FV^{i}_{M}]\\
  FV^{i}_{m} = AF^{i}_{m} \cdot ISF_{m}
  \label{eq:af_isf}
\end{align}
where $i$ and $M$ denote the image index and the number of actions, respectively. 
%
$AF^{i}_{m}$ is the frequency of $m$-th action class in each image.
$ISF_{M}$ is the inverse frequency of each action class in the dataset.
%
In AF-ISF, the similarity between $\bm{FV}^{j}$ and $\bm{FV}^{k}$, where $j$ and $k$ denote the IDs of query and retrieved images, evaluates the similarity of action structure of the two images.
%
If the cosine similarity exceeds a predefined threshold (e.g., 0.5), the query and retrieved images are regarded as matched as with the Action IoU.

While AF-ISF alleviates the action class imbalanced problem, several action classes are distinctive for representing a group scene even if the action distribution is balanced (e.g., ``Spiking'' is more important than ``Falling'' in VBD).
Due to this problem, AF-ISF is still improper for the contextual representation of group scenes.

\noindent{\bf Group activity retrieval.}
Based on the discussion above, we propose to further evaluate whether or not the group activity class of a query scene matches that of the retrieved scene.
Note that the group activity annotations are given to the test data (i.e., all query and retrieved images) only for this evaluation and are not used in our GAF learning. 

\subsubsection{Evaluation Metrics}
\label{subsubsection:evaluation_metrics}

With IoU and AF-ISF in the action set retrieval and group activity matching in the group activity retrieval, we compute the Hit@K 
used in~\cite{DBLP:conf/eccv/IbrahimM18}.
In addition, the mean Average Precision (mAP) is used in the action set retrieval as with~\cite{DBLP:conf/eccv/IbrahimM18}.
For mAP, the Euclidean distance of $\bm{G}$ between query and retrieved images is used as the confidence indicator.

\subsection{Training Details}

Our network is optimized by Adam~\cite{DBLP:journals/corr/KingmaB14} with $\beta_{1}=0.9$, $\beta_{2}=0.999$, and $\epsilon=10^{-8}$. 
%
%
The whole image is resized into 320x640 and 240x360 for VBD and CAD, respectively.
%
We employ the VGG-19 and Inception-v3 models as a person feature extractor (Fig.~\ref{fig:overview_network} (a)) for VBD and CAD, respectively.
%
While the person feature extractor trained on person action recognition is fine-tuned through our GAF learning in GAFL-PAC, we freeze the person feature extractor trained on ImageNet in GAFL-PAF following the previous method~\cite{DBLP:conf/eccv/IbrahimM18}.
Our APN consists of three fully-connected layers. 
%
%
As for the other details, we follow the widely used setting~\cite{DBLP:conf/iccv/Yuan0W21}. See the details in the supplementary material.

\subsection{Comparative Experiments}
\label{subsection:comp_exp}

We compare our methods with other methods by the following three types of experimental results: 
(1) the results of retrieval (Sec.~\ref{subsubsection:comp_ret}), 
(2) the results of GAR (Sec.~\ref{subsubsection:comp_gar}), and 
(3) the visualized distributions of GAFs
(Sec.~\ref{subsubsection:comp_tsne}).

\subsubsection{Retrieval}
\label{subsubsection:comp_ret}

To validate the effectiveness of the GAF for group representation, the set of person features (i.e., the concatenation of $\bm{F}_{grp}^{TS}$ and $\bm{F}_{grp}^{ST}$ in Fig.~\ref{fig:overview_network} (b)) is also evaluated.
Specifically, the output features of $P^{TM}$ in the TS and ST branches ($\bm{F}^{TS}_{grp}$ and $\bm{F}^{ST}_{grp}$ in Fig.~\ref{fig:overview_network}) are concatenated and also used for retrieval in our method.
%
Results obtained by the concatenation of output features (i.e., $\bm{F}^{TS}_{grp}$ and $\bm{F}^{ST}_{grp}$) and $\bm{G}$ are denoted as ``Ours-ind'' and ``Ours-grp'', respectively.
%

\noindent{\bf Volleyball dataset (GAFL-PAC).}
Our method is compared with the SOTA GAR methods~\cite{DBLP:conf/cvpr/0002Z0Y0CQ22,DBLP:conf/iccv/Yuan0W21,DBLP:journals/pami/YanXTST23} which are only trained with person action labels as with our method.
Table~\ref{table:comparison_volley} (top) shows that our method is best in action set and group activity retrieval. 
The results validate that our method learns features about the action structure and the group activity of a scene into the compact latent vector (i.e., $\bm{G}$) efficiently.
The large gain in the group activity retrieval shows that the features of multi-person activity are learned in our GAF in contrast to the SOTA GAR methods~\cite{DBLP:conf/cvpr/0002Z0Y0CQ22,DBLP:conf/iccv/Yuan0W21,DBLP:journals/pami/YanXTST23} where the set of person features is used for the retrieval.

\noindent{\bf Volleyball dataset (GAFL-PAF).}
Our method is compared with the SOTA method~\cite{DBLP:conf/eccv/IbrahimM18}, the only existing GAF learning method using no person-action and group-activity annotations, and baseline methods as with~\cite{DBLP:conf/eccv/IbrahimM18}.
From Table~\ref{table:comparison_volley} (bottom), we see that our method performs best in all metrics except that ``HRN'' is better than our method in mAP of Action set (IoU).
However, the performance difference is small (i.e., 46.9 and 46.4 in ``HRN'' and ``Ours-grp,'' respectively).
Furthermore, our method is better than ``HRN'' in Action set (AF-IDF).
From these results, we can say that our GAF is better even in the action set retrieval.
Regarding group activity retrieval, our method significantly outperforms the SOTA method. The results validate that our method learns the contextual features of multi-person activity better than the SOTA method.

\noindent{\bf Collective activity dataset (GAFL-PAC).}
The results in GAFL-PAC on CAD are shown in Table~\ref{table:comparison_cad} (top).
The results validate that our method is better than Dual-AI~\cite{DBLP:conf/cvpr/0002Z0Y0CQ22} in all metrics.
While ``Ours-grp'' is better than ``Ours-ind'' in most metrics, ``Ours-ind'' is better in mAP of Action set (IoU).
The action set (IoU) evaluates the similarity of the number of people, so the results may come from that the change in the number of people between scenes on this dataset is addressed well in ``Ours-ind.''
Specifically, we can consider that such information about the number of people may be preserved well in ``Ours-ind.'' where the set of person features is used for retrieval.

\noindent{\bf Collective activity dataset (GAFL-PAF).}
Table~\ref{table:comparison_cad} (bottom) shows that our method is the best in all metrics among the other methods.
%
The results validate the wide applicability of GAF learned even in general scenes included in CAD.
%
While ``Ours-grp'' is better than ``Ours-ind'' on VBD, ``Ours-ind'' is better on CAD.
%
These opposite results may come from the difference between the number of people in each image.
%
While $N=12$ people are observed in most images in VBD, around five people on average on CAD.
As the number of features in $\bm{F}^{TS}_{grp}$ and $\bm{F}^{ST}_{grp}$ increase in proportion to the number of observed people, the difficulty in learning $\bm{F}^{TS}_{grp}$ and $\bm{F}^{ST}_{grp}$ becomes higher.
%
Since this learning difficulty might occur on VBD, spatial max pooling used in ``Ours-grp'' is effective for reducing the feature dimension from $NC$ of $\bm{F}^{TS}_{grp}$ and $\bm{F}^{ST}_{grp}$ to $C$ of $\bm{G}^{TS}$ and $\bm{G}^{ST}$.

\begin{figure}[t]
  \begin{center}
  \includegraphics[width=\columnwidth]{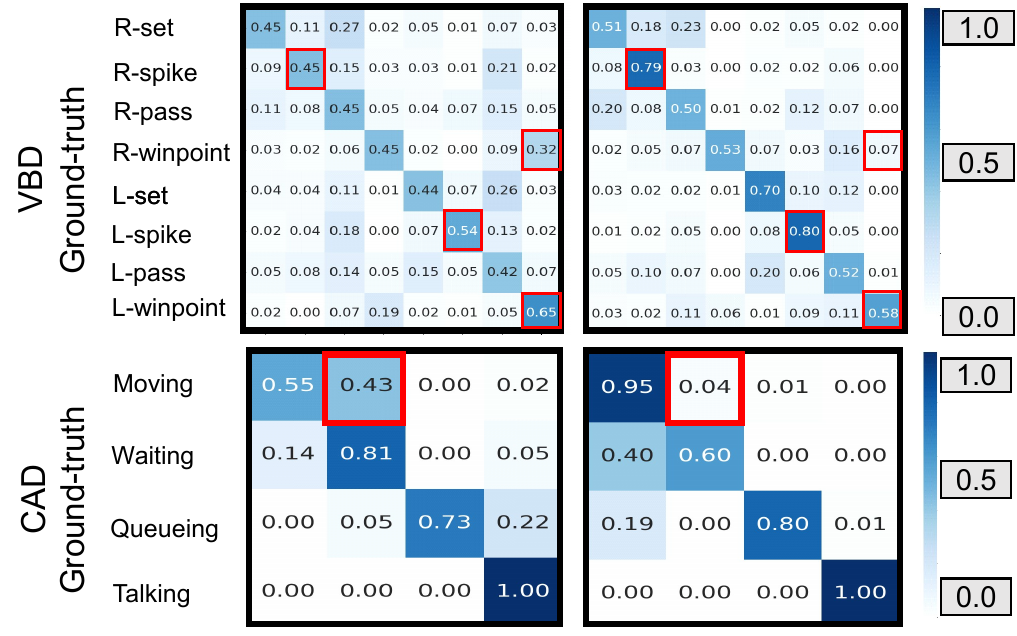}\\
  \vspace{-1mm}
  {\small
  ~\hspace*{3mm}
  B2-VGG19~\hspace{15mm}
  Ours
  }
  \end{center}
  \vspace{-4mm}
  \caption{
  Confusion matrices of GAR by nearest neighbor retrieval on VBD and CAD in GAFL-PAF.
  Each row and column show the ground-truth and recognized group activity, respectively. Results of the other methods are shown in the supplementary material.
  }
  \label{fig:cm_usl_all}
\end{figure}

\subsubsection{Group Activity Recognition}
\label{subsubsection:comp_gar}

While no group annotation is used in our GAF learning, the group activity class retrieved by Hit@1, which is equal to 1-nearest neighbor classification, can be regarded as the result of GAR, as done in~\cite{DBLP:conf/eccv/NakataNMMLD22}.
While results only in GAFL-PAF are shown in this section, results in GAFL-PAC are available in the supplementary material.

\noindent{\bf Volleyball dataset (GAFL-PAF).}
As shown in Fig.~\ref{fig:cm_usl_all}, our method outperforms ``B2-VGG19'' in all group activity classes except for L-winpoint.
While ``B2-VGG19'' is better than ``Ours'' in L-winpoint shown at the bottom right of each confusion matrix, ``B2-VGG19'' often misrecognizes the R-winpoint scene as L-winpoint, as shown in the red rectangle cells.
Left-spike and Right-spike are especially recognized better in our method than the others.
This superiority of our method can be interpreted as follows.
In Left-spike and Right-spike, spiking and blocking players who are distinctive both in appearance and location cues are always observed.
Since these distinctive people are important in predicting not only their attributes but also the attributes of other people, the features of these distinctive people are extracted well even in the compact latent vector (i.e., GAF).

\noindent{\bf Collective activity dataset (GAFL-PAF).}
The confusion matrices are shown in Fig.~\ref{fig:cm_usl_all}.
The results show that our method is the best in all activity classes.
In particular, while VGG19 gets many false negatives in Moving, the number of false negatives in Moving of our method is almost zero.

\begin{figure}[t]
  \begin{center}
  \includegraphics[width=\columnwidth]{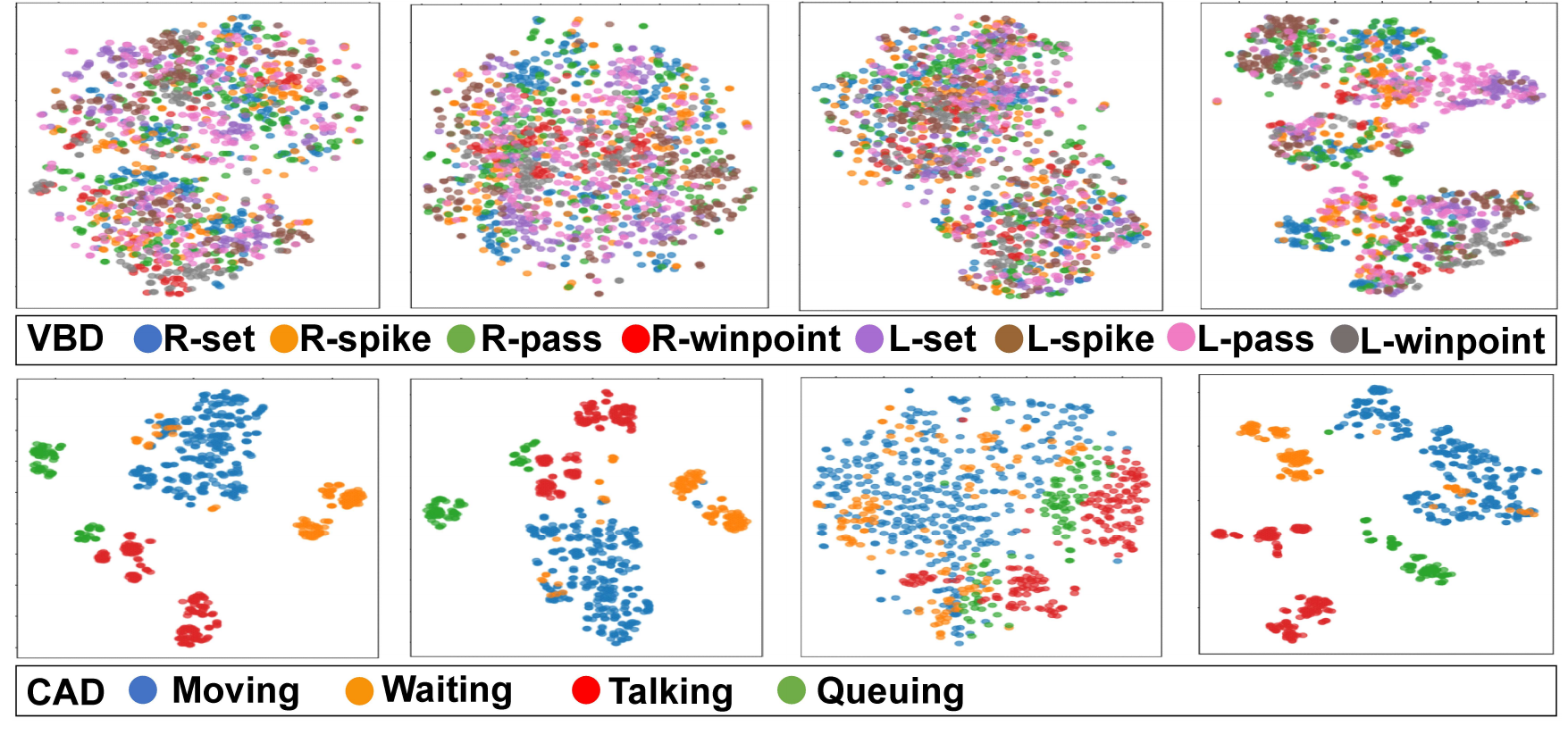}\\
  \vspace*{-2mm}
  \hspace*{-4mm}
  {\footnotesize
    B1-Compact128~\hspace{4mm}
    B2-VGG19~\hspace{10mm}
    HRN~\hspace{14mm}
    Ours~\hspace{3mm}
  }
  \end{center}
  \vspace{-4mm}
  \caption{Visualization of the learned GAF 
  on VBD and CAD in GAFL-PAF.
  The color of each sample shows the ground-truth of the group activity label corresponding to each test sample. 
  }
  \label{fig:latent_all_usl}
\end{figure}

\begin{figure}[t]
  \begin{center}
  \includegraphics[width=\columnwidth]{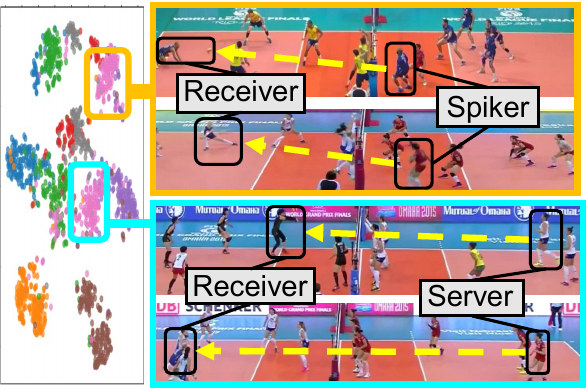}\\
  \end{center}
  \vspace{-4mm}
  \caption{Visualization of the learned GAF 
  on VBD in GAFL-PAC. 
  The magenta data points (i.e., ``L-pass'') are divided into two sub-categories based on the context (i.e., whether the receiving is caused by the spiking or serving of the opposite player).
  }
  \label{fig:latent_sl_volley_examples}
\end{figure}

\subsubsection{Visualization of Learned GAFs}
\label{subsubsection:comp_tsne}

The distribution of learned GAFs in all test images is visualized in a 2D space by t-SNE~\cite{van2008visualizing}.
The color of each point shows its annotated group activity class.
Figure~\ref{fig:latent_all_usl} shows that our method can learn the GAFs better than the other methods because (1) while the inner-class variance is small, the inter-class variance is large, and (2) the data points of similar group activities (e.g., Left-pass and Left-set) are closer.
 
Our GAFs in GAL-PAC on VBD are also shown in Fig.~\ref{fig:latent_sl_volley_examples}. 
This figure shows that data points with the same group activity labels are divided into sub-activities.
For example, L-pass indicated by magenta data points are divided into two clusters.
While the cluster enclosed by the orange rectangle represents L-pass where the person is receiving a ball from the opposite spiker, the other cluster enclosed by the light blue rectangle represents L-pass where the person is receiving a ball from the opposite server.
%
As shown in this example, our GAFs are learned well enough to represent visually subtle but important differences that are not represented in the manually defined activity classes.

\begin{figure}[t]
  \begin{center}
  \includegraphics[width=\columnwidth]{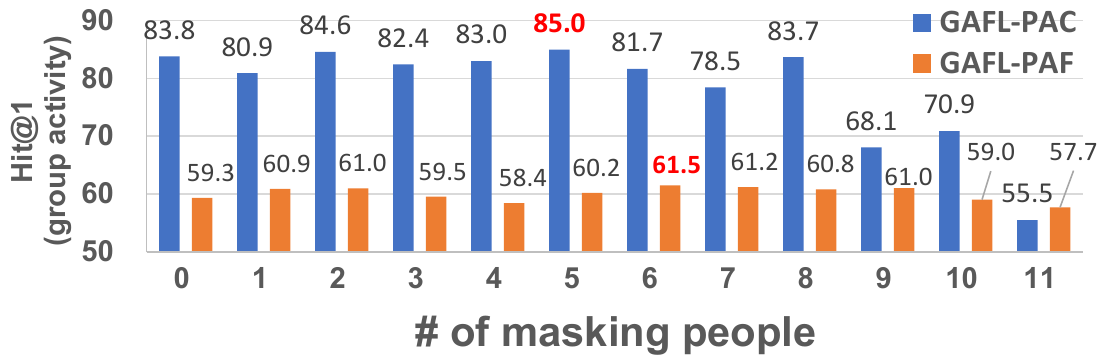}\\
  \end{center}
  \vspace{-3mm}
  \caption{
  Performance changes depending on the number of masking people on VBD.
  }
    \label{fig:mask_ratio_all_volley}
\end{figure}

\begin{table}[t]
    \centering
    \caption{Effectivenss of our location-guidance on VBD. 
    Results obtained by ``Ours-grp'' are shown as ``Ours.''
    }
    \vspace{-1mm}
    \begin{tabular}{l|l|c|c|c} \hline
    & \multicolumn{1}{c|}{\small{\shortstack{\\Retrieval type}}} & \multicolumn{1}{c|}{\footnotesize{\shortstack{\\Action set\\(IoU)}}} & \multicolumn{1}{c|}{\footnotesize{\shortstack{\\Action set\\(AF-IDF)}}} & \multicolumn{1}{c}{\footnotesize{\shortstack{\\Group\\activity}}}\\\cline{1-4}\hline
    & Method & Hit@1 & Hit@1 & Hit@1 \\ \hline\hline
    \multirow{2}{*}{\shortstack{GAFL-\\PAC}} & Ours w/o $\bm{F}_{loc}$ & 80.0 & 75.6 & 69.5 \\ \cline{2-5}
    & Ours & \red{83.0} & \red{80.1} & \red{84.8} \\ \hline\hline
    \multirow{2}{*}{\shortstack{GAFL-\\PAF}} & Ours w/o $\bm{F}_{loc}$ & 64.1 & 51.7 & 53.2 \\ \cline{2-5}
    & Ours & \red{64.8} & \red{52.3} & \red{61.1} \\ \hline
    \end{tabular}
    \label{table:det_abl_volley}
\end{table}

\subsection{Detailed analysis}
\label{subsection:det_analysis}

\noindent{\bf Comparison of the number of masked persons.}
We explore the optimal number of masked people (i.e., $N_{mask}$) for our MPM.
For this comparison, we change $N_{mask}$ from 0 to $N-1$. $N_{mask}=0$ means that $\bm{F}_{set}$ is directly fed into the transformer encoder without the MPM.

Figure~\ref{fig:mask_ratio_all_volley} shows that the performance changes depending on $N_{mask}$ on VBD.
The best performance is obtained when $N_{mask}$ is the middle number (i.e., 5 and 6 in the GAFL-PAC and GAFL-PAF, respectively), while the performance gain from $N_{mask}=0$ is insignificant.
%
%
%
On the other hand, we can also see that the performance with a large masking ratio (i.e., $N_{mask} \in \{10,11\}$) drops.
These results reveal that the extreme difficulty in the attribute prediction of the masked person from a few non-masked people leads GAF learning to failure.

\noindent{\bf Effect of location-guidance in GAF learning.}
%
We ablate $\bm{F}_{loc}^{p}$ which is used for guidance to extract the features of each person in the action prediction network in Fig.~\ref{fig:overview_network} (c) (see also Sec.~\ref{subsec:attribute_prediction}) on VBD.

In GAFL-PAC, ``Ours'' is better than ``Ours w/o $\bm{F}_{loc}^{p}$'' in the all metrics.
In particular, the performance gain in the group activity retrieval is large. 
%
We can interpret the reason as follows.
%
In ``Ours w/o $\bm{F}_{loc}^{p}$,'' only $\bm{G}$ is used to predict the attribute of each person in a scene. 
It makes the model predict not the attribute of each person but the attribute distribution of people in a scene.
%
Therefore, the GAF can be used for action set retrieval to some extent because the similarity of action distribution is evaluated in this retrieval.
However, spatial interaction between people is not learned in the GAF. 
It causes a significant performance drop in the group activity retrieval.
In GAFL-PAF, we can also see the high-performance gain of ``Ours'' in the group activity retrieval.
%
From these results, we can conclude that our location guidance is important for learning group activity.

\begin{table}[t]
    \centering
    \caption{
    Comparison with supervised GAR.
    Double lines separate the results obtained by VBD and CAD.
    }
    \vspace{-1mm}
    \begin{tabular}{c|l|c} \hline
    Dataset & Method & Accuracy \\ \hline\hline
    \multirow{2}{*}{\shortstack{VBD}} & Dual-AI~[\textcolor{green}{11}] & 92.1 \\ \cline{2-3}
    & Ours w/ group activity labels & \red{92.4} \\ \hline\hline
    \multirow{2}{*}{\shortstack{CAD}} & Dual-AI~[\textcolor{green}{11}] & 94.1 \\ \cline{2-3}
    & Ours w/ group activity labels & \red{96.6} \\ \hline
    \end{tabular}    
    \label{table:gar_sl_all}
\end{table}

\noindent{\bf Fine-tuning for Group Activity Recognition.}
Table~\ref{table:gar_sl_all} shows that the comparison of ``Dual-AI'' and ``Ours''
with group activity labels on VBD and CAD.
In ``Dual-AI,'' the network is trained with group activity labels from scratch.
In ``Ours,'' after the pretraining by our GAF learning without the group activity supervision, the network is fine-tuned for GAR with group activity labels.

On both datasets, our method is better than ``Dual-AI.'' 
In particular, on CAD, the GAR accuracy obtained by our method is 2.5$\%$ better than the GAR accuracy obtained by ``Dual-AI.''
The results demonstrate that our GAF learning is effective as a pre-training for supervised GAR.
\section{Concluding Remarks}
\label{sec:conclusion}

%
Instead of group activity annotations which is difficult due to a variety of similar group activities, our method learns GAF through person attribute prediction without group activity annotations.
Quantitative comparisons and visualized results show that our method can learn informative GAF compared with other methods on the two public datasets.

While our method outperforms all the other methods in our experiments not only for GAF learning but also for GAR, our method is understandably inferior to GAR supervised by group activity annotations.
%
Exploring other pretext tasks such as predicting the joint attention of a group (e.g.,~\cite{DBLP:conf/iccv/NakataniKU23}) is important future work for further GAF enhancement.
{
    \small
    \bibliographystyle{ieeenat_fullname}
    \bibliography{main}

\begin{thebibliography}{45}
\providecommand{\natexlab}[1]{#1}
\providecommand{\url}[1]{\texttt{#1}}
\expandafter\ifx\csname urlstyle\endcsname\relax
  \providecommand{\doi}[1]{doi: #1}\else
  \providecommand{\doi}{doi: \begingroup \urlstyle{rm}\Url}\fi

\bibitem[Aizawa(2003)]{DBLP:journals/ipm/Aizawa03}
Akiko~N. Aizawa.
\newblock An information-theoretic perspective of tf-idf measures.
\newblock \emph{Inf. Process. Manag.}, 39\penalty0 (1):\penalty0 45--65, 2003.

\bibitem[Azar et~al.(2019)Azar, Atigh, Nickabadi, and Alahi]{DBLP:conf/cvpr/AzarANA19}
Sina~Mokhtarzadeh Azar, Mina~Ghadimi Atigh, Ahmad Nickabadi, and Alexandre Alahi.
\newblock Convolutional relational machine for group activity recognition.
\newblock In \emph{CVPR}, 2019.

\bibitem[Choi et~al.(2009)Choi, Shahid, and Savarese]{DBLP:conf/iccvw/ChoiSS09}
Wongun Choi, Khuram Shahid, and Silvio Savarese.
\newblock What are they doing? : Collective activity classification using spatio-temporal relationship among people.
\newblock In \emph{ICCVW}, 2009.

\bibitem[Doersch et~al.(2015)Doersch, Gupta, and Efros]{DBLP:conf/iccv/DoerschGE15}
Carl Doersch, Abhinav Gupta, and Alexei~A. Efros.
\newblock Unsupervised visual representation learning by context prediction.
\newblock In \emph{ICCV}, 2015.

\bibitem[Duan et~al.(2022)Duan, Zhao, Chen, and Lin]{DBLP:conf/cvpr/DuanZCL22}
Haodong Duan, Nanxuan Zhao, Kai Chen, and Dahua Lin.
\newblock Transrank: Self-supervised video representation learning via ranking-based transformation recognition.
\newblock In \emph{CVPR}, 2022.

\bibitem[Ehsanpour et~al.(2020)Ehsanpour, Abedin, Saleh, Shi, Reid, and Rezatofighi]{DBLP:conf/eccv/EhsanpourASSRR20}
Mahsa Ehsanpour, Alireza Abedin, Fatemeh~Sadat Saleh, Javen Shi, Ian~D. Reid, and Hamid Rezatofighi.
\newblock Joint learning of social groups, individuals action and sub-group activities in videos.
\newblock In \emph{ECCV}, 2020.

\bibitem[Fernando et~al.(2017)Fernando, Bilen, Gavves, and Gould]{DBLP:conf/cvpr/FernandoBGG17}
Basura Fernando, Hakan Bilen, Efstratios Gavves, and Stephen Gould.
\newblock Self-supervised video representation learning with odd-one-out networks.
\newblock In \emph{CVPR}, 2017.

\bibitem[Foster(2016)]{bib:AFP2016}
Richard~B. Foster.
\newblock \emph{American Football Playbook: 210 Field Templates}.
\newblock Createspace Independent Pub, 2016.

\bibitem[Gavrilyuk et~al.(2020)Gavrilyuk, Sanford, Javan, and Snoek]{DBLP:conf/cvpr/GavrilyukSJS20}
Kirill Gavrilyuk, Ryan Sanford, Mehrsan Javan, and Cees G.~M. Snoek.
\newblock Actor-transformers for group activity recognition.
\newblock In \emph{CVPR}, 2020.

\bibitem[Gidaris et~al.(2018)Gidaris, Singh, and Komodakis]{DBLP:conf/iclr/GidarisSK18}
Spyros Gidaris, Praveer Singh, and Nikos Komodakis.
\newblock Unsupervised representation learning by predicting image rotations.
\newblock In \emph{ICLR}, 2018.

\bibitem[Han et~al.(2022)Han, Zhang, Wang, Yan, Yao, Chang, and Qiao]{DBLP:conf/cvpr/0002Z0Y0CQ22}
Mingfei Han, David~Junhao Zhang, Yali Wang, Rui Yan, Lina Yao, Xiaojun Chang, and Yu Qiao.
\newblock Dual-ai: Dual-path actor interaction learning for group activity recognition.
\newblock In \emph{CVPR}, 2022.

\bibitem[He et~al.(2017)He, Gkioxari, Doll{\'{a}}r, and Girshick]{DBLP:conf/iccv/HeGDG17}
Kaiming He, Georgia Gkioxari, Piotr Doll{\'{a}}r, and Ross~B. Girshick.
\newblock Mask {R-CNN}.
\newblock In \emph{ICCV}, 2017.

\bibitem[Heilbron et~al.(2018)Heilbron, Lee, Jin, and Ghanem]{DBLP:conf/eccv/HeilbronLJG18}
Fabian~Caba Heilbron, Joon{-}Young Lee, Hailin Jin, and Bernard Ghanem.
\newblock What do {I} annotate next? an empirical study of active learning for action localization.
\newblock In \emph{ECCV}, 2018.

\bibitem[Ibrahim and Mori(2018)]{DBLP:conf/eccv/IbrahimM18}
Mostafa~S. Ibrahim and Greg Mori.
\newblock Hierarchical relational networks for group activity recognition and retrieval.
\newblock In \emph{ECCV}, 2018.

\bibitem[Ibrahim et~al.(2016)Ibrahim, Muralidharan, Deng, Vahdat, and Mori]{DBLP:conf/cvpr/IbrahimMDVM16}
Mostafa~S. Ibrahim, Srikanth Muralidharan, Zhiwei Deng, Arash Vahdat, and Greg Mori.
\newblock A hierarchical deep temporal model for group activity recognition.
\newblock In \emph{CVPR}, 2016.

\bibitem[Kim et~al.(2022)Kim, Lee, Cho, and Kwak]{DBLP:conf/cvpr/KimLCK22}
Dongkeun Kim, Jinsung Lee, Minsu Cho, and Suha Kwak.
\newblock Detector-free weakly supervised group activity recognition.
\newblock In \emph{CVPR}, 2022.

\bibitem[Kingma and Ba(2015)]{DBLP:journals/corr/KingmaB14}
Diederik~P. Kingma and Jimmy Ba.
\newblock Adam: {A} method for stochastic optimization.
\newblock In \emph{ICLR}, 2015.

\bibitem[Larsson et~al.(2017)Larsson, Maire, and Shakhnarovich]{DBLP:conf/cvpr/LarssonMS17}
Gustav Larsson, Michael Maire, and Gregory Shakhnarovich.
\newblock Colorization as a proxy task for visual understanding.
\newblock In \emph{CVPR}, 2017.

\bibitem[Li et~al.(2021)Li, Cao, Liu, Yang, Liu, Hou, and Yi]{DBLP:conf/iccv/LiCLYLHY21}
Shuaicheng Li, Qianggang Cao, Lingbo Liu, Kunlin Yang, Shinan Liu, Jun Hou, and Shuai Yi.
\newblock Groupformer: Group activity recognition with clustered spatial-temporal transformer.
\newblock In \emph{ICCV}, 2021.

\bibitem[Nakata et~al.(2022)Nakata, Ng, Miyashita, Maki, Lin, and Deguchi]{DBLP:conf/eccv/NakataNMMLD22}
Kengo Nakata, Youyang Ng, Daisuke Miyashita, Asuka Maki, Yu{-}Chieh Lin, and Jun Deguchi.
\newblock Revisiting a knn-based image classification system with high-capacity storage.
\newblock In \emph{ECCV}, 2022.

\bibitem[Nakatani et~al.(2021)Nakatani, Sendo, and Ukita]{DBLP:conf/mva/NakataniSU21}
Chihiro Nakatani, Kohei Sendo, and Norimichi Ukita.
\newblock Group activity recognition using joint learning of individual action recognition and people grouping.
\newblock In \emph{MVA}, 2021.

\bibitem[Nakatani et~al.(2023)Nakatani, Kawashima, and Ukita]{DBLP:conf/iccv/NakataniKU23}
Chihiro Nakatani, Hiroaki Kawashima, and Norimichi Ukita.
\newblock Interaction-aware joint attention estimation using people attributes.
\newblock In \emph{ICCV}, 2023.

\bibitem[Noroozi and Favaro(2016)]{DBLP:conf/eccv/NorooziF16}
Mehdi Noroozi and Paolo Favaro.
\newblock Unsupervised learning of visual representations by solving jigsaw puzzles.
\newblock In \emph{ECCV}, 2016.

\bibitem[Pathak et~al.(2016)Pathak, Kr{\"{a}}henb{\"{u}}hl, Donahue, Darrell, and Efros]{DBLP:conf/cvpr/PathakKDDE16}
Deepak Pathak, Philipp Kr{\"{a}}henb{\"{u}}hl, Jeff Donahue, Trevor Darrell, and Alexei~A. Efros.
\newblock Context encoders: Feature learning by inpainting.
\newblock In \emph{CVPR}, 2016.

\bibitem[Pramono et~al.(2020)Pramono, Chen, and Fang]{DBLP:conf/eccv/PramonoCF20}
Rizard Renanda~Adhi Pramono, Yie{-}Tarng Chen, and Wen{-}Hsien Fang.
\newblock Empowering relational network by self-attention augmented conditional random fields for group activity recognition.
\newblock In \emph{ECCV}, 2020.

\bibitem[Rana and Rawat(2023)]{DBLP:conf/cvpr/RanaR23}
Aayush~Jung Rana and Yogesh~S. Rawat.
\newblock Hybrid active learning via deep clustering for video action detection.
\newblock In \emph{CVPR}, 2023.

\bibitem[Simonyan and Zisserman(2015)]{DBLP:journals/corr/SimonyanZ14a}
Karen Simonyan and Andrew Zisserman.
\newblock Very deep convolutional networks for large-scale image recognition.
\newblock In \emph{ICLR}, 2015.

\bibitem[Sultani et~al.(2018)Sultani, Chen, and Shah]{DBLP:conf/cvpr/SultaniCS18}
Waqas Sultani, Chen Chen, and Mubarak Shah.
\newblock Real-world anomaly detection in surveillance videos.
\newblock In \emph{CVPR}, 2018.

\bibitem[Tamura et~al.(2022)Tamura, Vishwakarma, and Vennelakanti]{DBLP:conf/eccv/TamuraVV22}
Masato Tamura, Rahul Vishwakarma, and Ravigopal Vennelakanti.
\newblock Hunting group clues with transformers for social group activity recognition.
\newblock In \emph{ECCV}, 2022.

\bibitem[Van~der Maaten and Hinton(2008)]{van2008visualizing}
Laurens Van~der Maaten and Geoffrey Hinton.
\newblock Visualizing data using t-sne.
\newblock \emph{Journal of machine learning research}, 9\penalty0 (11), 2008.

\bibitem[Vaswani et~al.(2017)Vaswani, Shazeer, Parmar, Uszkoreit, Jones, Gomez, Kaiser, and Polosukhin]{DBLP:conf/nips/VaswaniSPUJGKP17}
Ashish Vaswani, Noam Shazeer, Niki Parmar, Jakob Uszkoreit, Llion Jones, Aidan~N. Gomez, Lukasz Kaiser, and Illia Polosukhin.
\newblock Attention is all you need.
\newblock In \emph{NIPS}, 2017.

\bibitem[Wang et~al.(2019)Wang, Jiao, Bao, He, Liu, and Liu]{DBLP:conf/cvpr/WangJBHLL19}
Jiangliu Wang, Jianbo Jiao, Linchao Bao, Shengfeng He, Yunhui Liu, and Wei Liu.
\newblock Self-supervised spatio-temporal representation learning for videos by predicting motion and appearance statistics.
\newblock In \emph{CVPR}, 2019.

\bibitem[Wang et~al.(2020)Wang, Jiao, and Liu]{DBLP:conf/eccv/WangJL20}
Jiangliu Wang, Jianbo Jiao, and Yun{-}Hui Liu.
\newblock Self-supervised video representation learning by pace prediction.
\newblock In \emph{ECCV}, 2020.

\bibitem[Wang et~al.(2017)Wang, Ni, and Yang]{DBLP:conf/cvpr/WangNY17}
Minsi Wang, Bingbing Ni, and Xiaokang Yang.
\newblock Recurrent modeling of interaction context for collective activity recognition.
\newblock In \emph{CVPR}, 2017.

\bibitem[Wang et~al.(2022)Wang, Zhang, Qing, Tang, Zuo, Gao, Jin, and Sang]{DBLP:conf/cvpr/WangZQTZG0S22}
Xiang Wang, Shiwei Zhang, Zhiwu Qing, Mingqian Tang, Zhengrong Zuo, Changxin Gao, Rong Jin, and Nong Sang.
\newblock Hybrid relation guided set matching for few-shot action recognition.
\newblock In \emph{CVPR}, 2022.

\bibitem[Wu et~al.(2022)Wu, Zhao, Bao, and Wildes]{DBLP:conf/eccv/WuZBW22}
Dekun Wu, He Zhao, Xingce Bao, and Richard~P. Wildes.
\newblock Sports video analysis on large-scale data.
\newblock In \emph{ECCV}, 2022.

\bibitem[Wu et~al.(2019)Wu, Wang, Wang, Guo, and Wu]{DBLP:conf/cvpr/WuWWGW19}
Jianchao Wu, Limin Wang, Li Wang, Jie Guo, and Gangshan Wu.
\newblock Learning actor relation graphs for group activity recognition.
\newblock In \emph{CVPR}, 2019.

\bibitem[Xie et~al.(2023)Xie, Gao, Wu, and Chang]{DBLP:conf/cvpr/XieGWC23}
Zhao Xie, Tian Gao, Kewei Wu, and Jiao Chang.
\newblock An actor-centric causality graph for asynchronous temporal inference in group activity.
\newblock In \emph{CVPR}, 2023.

\bibitem[Yan et~al.(2020)Yan, Xie, Tang, Shu, and Tian]{DBLP:conf/eccv/YanXTS020}
Rui Yan, Lingxi Xie, Jinhui Tang, Xiangbo Shu, and Qi Tian.
\newblock Social adaptive module for weakly-supervised group activity recognition.
\newblock In \emph{ECCV}, 2020.

\bibitem[Yan et~al.(2023)Yan, Xie, Tang, Shu, and Tian]{DBLP:journals/pami/YanXTST23}
Rui Yan, Lingxi Xie, Jinhui Tang, Xiangbo Shu, and Qi Tian.
\newblock Higcin: Hierarchical graph-based cross inference network for group activity recognition.
\newblock \emph{{IEEE} Trans. Pattern Anal. Mach. Intell.}, 45\penalty0 (6):\penalty0 6955--6968, 2023.

\bibitem[Yao et~al.(2020)Yao, Liu, Luo, Zhou, and Ye]{DBLP:conf/cvpr/YaoLLZY20}
Yuan Yao, Chang Liu, Dezhao Luo, Yu Zhou, and Qixiang Ye.
\newblock Video playback rate perception for self-supervised spatio-temporal representation learning.
\newblock In \emph{CVPR}, 2020.

\bibitem[Yuan et~al.(2021)Yuan, Ni, and Wang]{DBLP:conf/iccv/Yuan0W21}
Hangjie Yuan, Dong Ni, and Mang Wang.
\newblock Spatio-temporal dynamic inference network for group activity recognition.
\newblock In \emph{ICCV}, 2021.

\bibitem[Zhang et~al.(2019)Zhang, Qi, Wang, and Luo]{DBLP:conf/cvpr/ZhangQWL19}
Liheng Zhang, Guo{-}Jun Qi, Liqiang Wang, and Jiebo Luo.
\newblock {AET} vs. {AED:} unsupervised representation learning by auto-encoding transformations rather than data.
\newblock In \emph{CVPR}, 2019.

\bibitem[Zheng et~al.(2022)Zheng, Chen, and Jin]{DBLP:conf/eccv/ZhengCJ22}
Sipeng Zheng, Shizhe Chen, and Qin Jin.
\newblock Few-shot action recognition with hierarchical matching and contrastive learning.
\newblock In \emph{ECCV}, 2022.

\bibitem[Zhou et~al.(2022)Zhou, Kadav, Shamsian, Geng, Lai, Zhao, Liu, Kapadia, and Graf]{DBLP:conf/eccv/ZhouKSGLZLKG22}
Honglu Zhou, Asim Kadav, Aviv Shamsian, Shijie Geng, Farley Lai, Long Zhao, Ting Liu, Mubbasir Kapadia, and Hans~Peter Graf.
\newblock {COMPOSER:} compositional reasoning of group activity in videos with keypoint-only modality.
\newblock In \emph{ECCV}, 2022.

\end{thebibliography}
}

\clearpage

\setcounter{figure}{7}
\setcounter{table}{4}

\section{Implementation Details}
\label{sec:implementaion_details}

This section shows the implementation details that are not mentioned in the main paper.

\subsection{Experimental Conditions}
\label{subsec:trainings}

We trained all of our models on the Nvidia A100 GPUs with batch size 8. 
For pertaining the person feature extractor in the GAFL-PAC, we set the learning rates to be 0.0001 and 0.00005 for the volleyball and collective activity datasets, respectively.
As mentioned in the main paper, the person feature extractor is fine-tuned through our GAF learning in GAFL-PAC while we freeze the person feature extractor trained on ImageNet in GAFL-PAF following the previous method~[\textcolor{green}{14}]. 
As the person feature extractor, we utilize the code implemented for DIN~[\textcolor{green}{39}] available at \url{https://github.com/JacobYuan7/DIN-Group-Activity-Recognition-Benchmark}.
In our GAF learning, the learning rates are 0.0001 and 0.00005 for the volleyball and collective activity datasets, respectively.

We utilize the experimental setting of DIN~[\textcolor{green}{39}] as follows.
We use video clips consisting of 10 frames (i.e., $T=10$) for both the volleyball and collective activity datasets.
The feature of each person is embedded into 1024 dimensional vectors (i.e., $C=1024$) after RoIAlign in the person feature extractor.
The RoIAlign in the person feature extractor is applied to each person with the ground-truth full-body bounding boxes as used in~[\textcolor{green}{11},\textcolor{green}{14},\textcolor{green}{37},\textcolor{green}{39}]. 
For evaluation, the threshold of action set retrieval (i.e., IoU and AF-IDF) is defined as 0.5 following ~[\textcolor{green}{14}].

\subsection{Implementation of Previous Methods}
\label{subsec:imple_pre_methods} 
All methods, including our method, are evaluated in the same setting for a fair comparison. 

\subsubsection{GAFL-PAC}
HiGCIN~[\textcolor{green}{37}], DIN~[\textcolor{green}{39}] and Dual-AI~[\textcolor{green}{11}] are used for a comparison in GAFL-PAC.
As mentioned in the main paper, we train these models with person action labels for a fair comparison as with our method in GAFL-PAC.
These codes are prepared as follows:
\begin{itemize}
    \item {\bf HiGCIN and DIN}: These codes are available at \url{https://github.com/JacobYuan7/DIN-Group-Activity-Recognition-Benchmark}. 
    \item {\bf Dual-AI}: The code is not available. We implement the method based on~[\textcolor{green}{11}].
\end{itemize}

\subsubsection{GAFL-PAF}
As with the GAFL-PAC, we compare our method with the SOTA methods in GAFL-PAF.
For evaluation, B1-Compact~[\textcolor{green}{14}], B2-VGG18~[\textcolor{green}{14}], and HRN~[\textcolor{green}{14}] are used for comparison because these methods are trained without group activity and person action annotations as with our method in GAFL-PAF.
These codes are prepared as follows:
\begin{itemize}
    \item {\bf B1-Compact}: The code is not available. However, we implement the method based on~[\textcolor{green}{14}] since the details of the model are mentioned in~[\textcolor{green}{14}].
    The dimension of each person feature is $128$.
    \item {\bf B2-VGG19}: As with B1-Compact, we implement B2-VGG19 based on~[\textcolor{green}{14}]. 
    In this method, the output of fc7 layer in a pretrained VGG19 network is regarded as a person feature of each person. 
    The dimension of each person feature is $4096$.
    \item {\bf HRN}: While the code is available at \url{https://github.com/mostafa-saad/hierarchical-relational-network}, the retrieval function is not included in the provided code. 
    Therefore, we implement the network for the retrieval function using person features acquired by HRN. 
    This network is based on~[\textcolor{green}{14}]. 
    The dimension of each person feature is $128$ as with B1-Compact.
\end{itemize}

\section{Additional Experiments}
\label{sec:add_exp_results}

Additional experiments, which are not included in the main paper for the page limitation, are presented in this section.
As with the main paper, comparative experiments and detailed analysis are shown in Secs.~\ref{subsection:comp_exp_supp} and~\ref{subsection:det_analysis_supp} in this supplementary material.

\subsection{Comparative Experiments}
\label{subsection:comp_exp_supp}

\begin{table*}[t]
    \centering
    \caption{
    Additional quantitative comparison of retrieval on the VolleyBall Dataset (VBD) in GAFL-PAF.
    In addition to the results shown in Table~\textcolor{red}{1} of the main paper, variants of these previous methods using $\bm{F}_{loc}$ are compared with our method in this table.
    The best result in each column is colored in \textcolor{red}{red}.
    Results obtained by the concatenation of output features (i.e., $\bm{F}^{TS}_{grp}$ and $\bm{F}^{ST}_{grp}$) and $\bm{G}$ are denoted as ``Ours-ind'' and ``Ours-grp'', respectively.
    }
    {\small
    \begin{tabular}{l||r|r|r|r|r|r|r|r|r|r|r} \hline
    Retrieval type & \multicolumn{4}{c|}{Action set (IoU~[\textcolor{green}{14}]} & \multicolumn{4}{c|}{Action set (AF-IDF)} & \multicolumn{3}{c}{Group activity}\\ \hline
    Method & Hit@1 & Hit@2 & Hit@3 & mAP & Hit@1 & Hit@2 & Hit@3 & mAP & Hit@1 & Hit@2 & Hit@3 \\\hline \hline
    B1-Compact128 & 57.9 & 75.7 & 84.3 & 45.8 & 41.3 & 60.8 & 71.4 & 29.3 & 30.3 & 48.0 & 59.9 \\ \hline
    B1-Compact128 w/ $\bm{F}_{loc}$ & 60.6 & 80.9 & 88.0 & 45.9 & 46.1 & 64.1 & 74.8 & 29.3 & 34.8 & 52.1 & 63.6 \\ \hline
    B2-VGG19 & 63.8 & 80.6 & 86.8 & 46.8 & 46.7 & 65.8 & 75.7 & 29.4 & 35.4 & 53.6 & 65.0 \\ \hline
    B2-VGG19 w/ $\bm{F}_{loc}$ & 63.6 & 80.5 & 88.0 & 46.1 & 48.5 & 66.6 & 76.3 & 28.9 & 51.6 & 69.1 & 78.3 \\ \hline
    HRN & 60.9 & 78.6 & 86.0 & \red{46.9} & 40.8 & 60.9 & 72.9 & 28.7 & 31.2 & 47.0 & 57.6 \\ \hline
    HRN w/ $\bm{F}_{loc}$ & 60.3 & 77.9 & 85.0 & \red{46.9} & 42.3 & 62.2 & 73.2 & 28.7 & 29.3 & 44.9 & 56.8 \\ \hline\hline
    Ours-ind & 64.2 & 80.8 & 88.3 & 45.0 & 50.4 & 69.3 & 77.6 & 30.1 & 55.0 & 72.3 & 79.2 \\ \hline
    Ours-grp & \red{64.8} & \red{82.7} & \red{90.3} & 46.4 & \red{52.3} & \red{71.4} & \red{81.0} & \red{31.4} & \red{61.1} & \red{75.1} & \red{82.4} \\ \hline
    \end{tabular}
    }
    \label{table:comparison_volley_add}
\end{table*}

\begin{table*}[t]
    \centering
    \caption{
    Additional quantitative comparison of retrieval on the Collective Activity Dataset (CAD) in GAFL-PAF.
    In addition to the results shown in Table~\textcolor{red}{2} of the main paper, variants of these previous methods using $\bm{F}_{loc}$ are compared with our method in this table.
    }
    {\small
    \begin{tabular}{l||r|r|r|r|r|r|r|r|r|r|r} \hline
    Retrieval type & \multicolumn{4}{c|}{Action set (IoU~[\textcolor{green}{14}]} & \multicolumn{4}{c|}{Action set (AF-IDF)} & \multicolumn{3}{c}{Group activity}\\ \hline
    Method & Hit@1 & Hit@2 & Hit@3 & mAP & Hit@1 & Hit@2 & Hit@3 & mAP & Hit@1 & Hit@2 & Hit@3 \\\hline\hline
    B1-Compact128 & 48.8 & 60.3 & 68.2 & 38.0 & 81.8 & 88.2 & 89.7 & 52.6 & 82.4 & 88.4 & 90.1 \\ \hline
    B1-Compact128 w/ $\bm{F}_{loc}$ & 59.1 & 73.7 & 78.7 & 40.1 & 70.7 & 79.1 & 82.2 & 46.4 & 70.7 & 79.6 & 82.9 \\ \hline
    B2-VGG19 & 53.6 & 61.6 & 66.1 & 35.3 & 71.1 & 80.3 & 83.8 & 46.7 & 72.2 & 80.8 & 84.2 \\ \hline
    B2-VGG19 w/ $\bm{F}_{loc}$ & 43.5 & 54.8 & 60.9 & 37.0 & 81.2 & 86.7 & 89.3 & 52.2 & 80.9 & 85.1 & 86.9 \\ \hline
    HRN & 37.1 & 50.1 & 58.6 & 22.2 & 53.2 & 64.8 & 72.5 & 34.2 & 54.0 & 64.8 & 72.4 \\ \hline
    HRN w/ $\bm{F}_{loc}$ & 32.0 & 45.8 & 54.2 & 21.1 & 49.2 & 60.7 & 69.8 & 32.9 & 49.9 & 60.7 & 69.3 \\ \hline\hline
    Ours-ind & \red{67.6} & \red{81.3} & \red{85.9} & \red{53.3} & \red{83.7} & \red{88.9} & \red{90.2} & 57.5 & \red{88.5} & \red{91.2} & \red{91.9} \\ \hline
    Ours-grp & 52.7 & 70.3 & 74.1 & 46.4 & 74.0 & 80.5 & 82.6 & \red{60.1} & 79.2 & 81.0 & 82.0 \\ \hline
    \end{tabular}
    }
    \label{table:comparison_cad_add}
\end{table*}

\subsubsection{Retrieval}
As noted in~[\textcolor{green}{14}], previous methods (i.e., B1-Compact, B2-VGG19, and HRN) sort person image features of all people based on their locations to avoid comparison with each possible permutation for retrieval in Sec.~\textcolor{red}{4.4.1} of the main paper.
However, the location-based sorting only captures the spatial interaction between people coarsely.
Therefore, we also compare variants of these previous methods in which location features (denoted by $\bm{F}_{loc}$ as with our method) are added to their person image features to capture such spatial interaction precisely.
As with our method, the location of each person is encoded in a feature vector by positional encoding.

\paragraph{Volleyball dataset (GAFL-PAF).}
Table~\ref{table:comparison_volley_add} shows the results obtained by the previous methods (i.e., equal to the results shown in Table~\textcolor{red}{1} of the main paper) and results obtained by their variants.
In both action set and group activity retrieval, we can see that adding $\bm{F}_{loc}$ is somewhat effective in these previous methods.
Specially, the performance gain between ``B2-VGG19'' and ``B2-VGG19 w/ $\bm{F}_{loc}$'' is large (i.e., 16.2 \%).
However, our method is still better than these variants in all metrics.
These results demonstrate that our GAF learned through location-guided person attribute prediction in the APN (Fig.~\textcolor{red}{3} (c) of the main paper) is better for representing people in a group.  

\paragraph{Collective activity dataset (GAFL-PAF).}
As with the volleyball dataset, the results on the collective activity dataset are shown in Table~\ref{table:comparison_cad_add}.
Different from the results on the volleyball dataset, $\bm{F}_{loc}$ has a negative impact on the collective activity dataset.
These results may come from the fact that $\bm{F}_{loc}$ makes the training of the person feature extractor complex in these previous methods.
For example, in B2-VGG19, we can see the positive impact of $\bm{F}_{loc}$.
This is because the method only employs the frozen model (i.e., VGG19) without any training.
Compared with all of these methods, our method is still the best in all metrics.
These results reveal that our location-guidance disentangles such location information in the complex GAF, as its effectiveness is also validated in Tables~\textcolor{red}{3} and~\ref{table:det_abl_cad} of the main paper and the supplementary materials.

\begin{figure*}[t]
  \begin{center}
  \includegraphics[width=\textwidth]{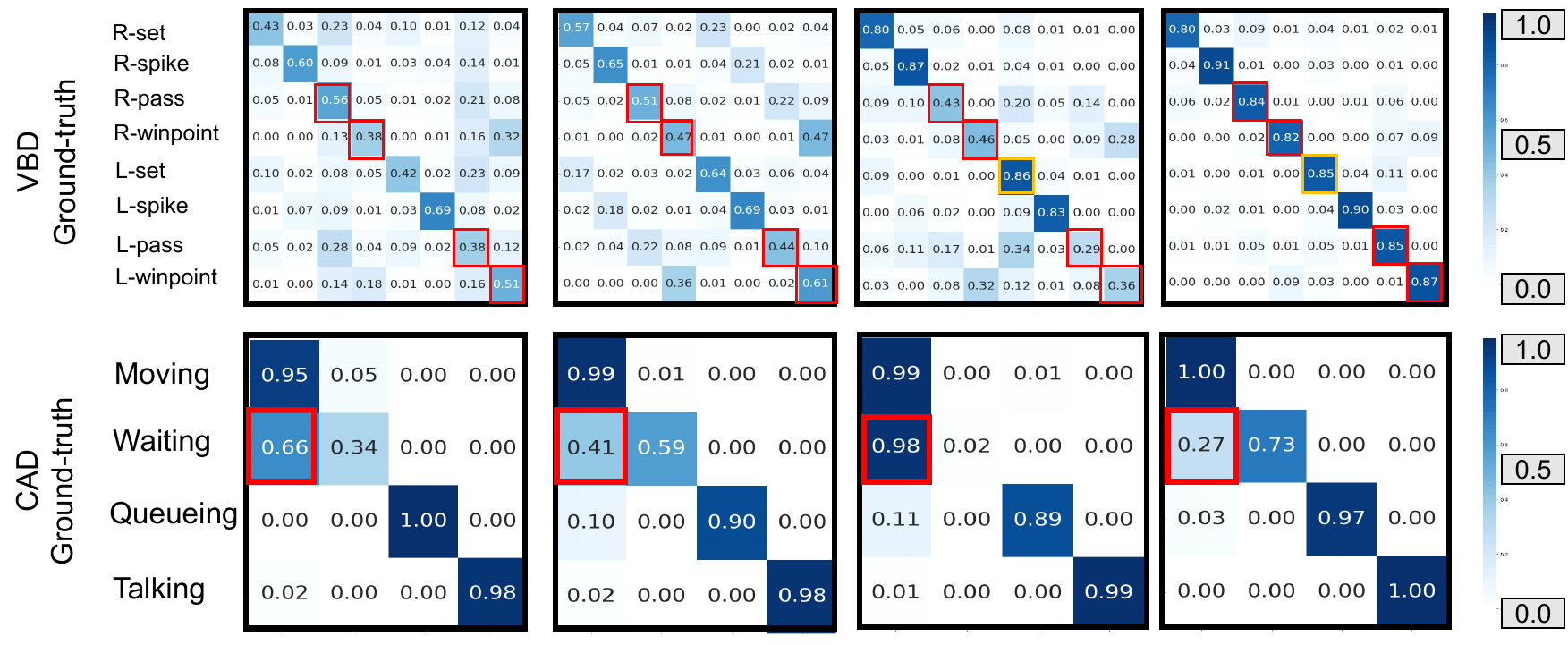}\\
  \vspace{-1mm}
  \hspace{6mm}
  HiGCIN~[\textcolor{green}{37}]~\hspace{17mm}
  DIN~[\textcolor{green}{39}]~\hspace{17mm}
  Dual-AI~[\textcolor{green}{11}]~\hspace{15mm}
  Ours
  \end{center}
  \vspace{-2mm}
  \caption{Confusion matrices of GAR by nearest neighbor retrieval on the VolleyBall Dataset (VBD) and Collective Activity Dataset (CAD) in GAFL-PAC.
  Each row and column show the ground-truth of the GA label and the recognized GA, respectively.
  }
  \label{fig:cm_sl_all}
\end{figure*}

\begin{figure*}[t]
  \begin{center}
  \includegraphics[width=\textwidth]{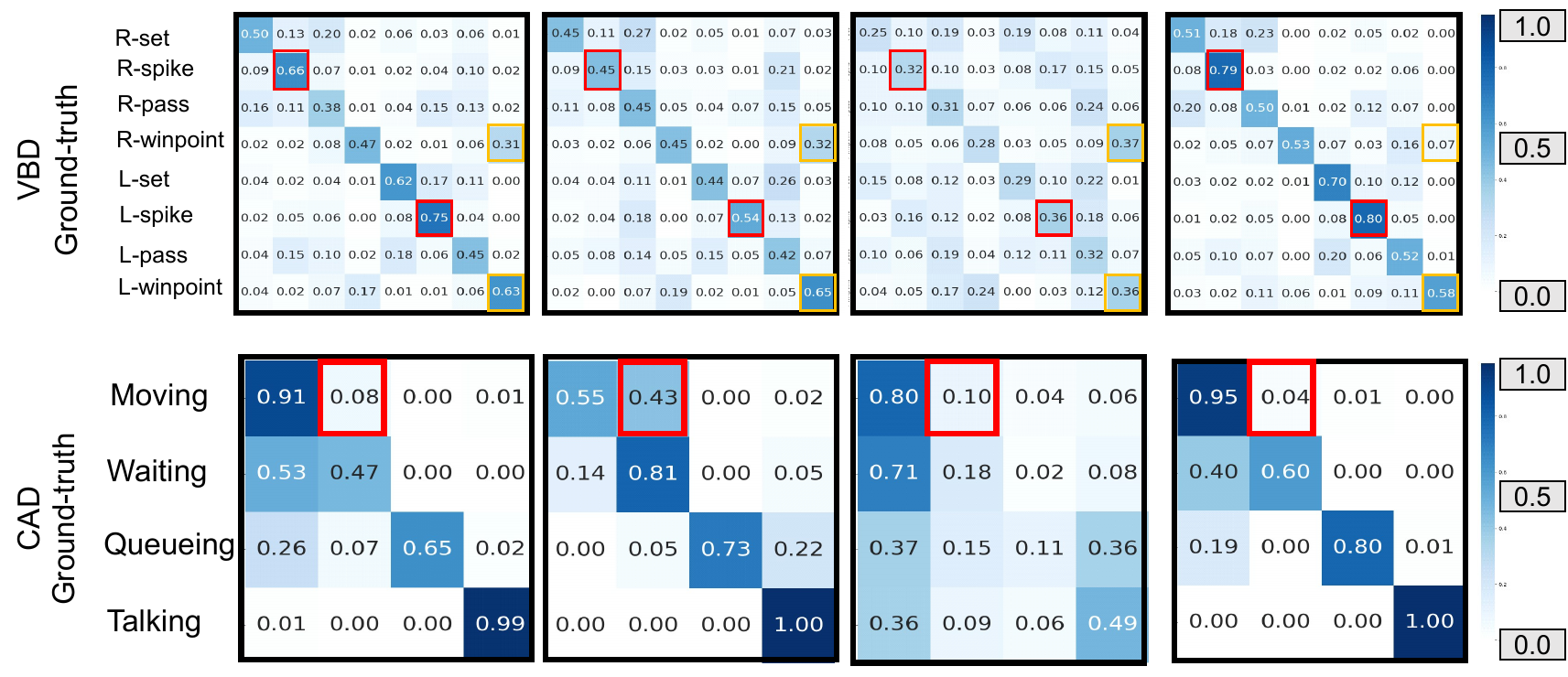}\\
  \vspace{-1mm}
  \hspace{5mm}
  B1-Compact128~[\textcolor{green}{14}]~\hspace{5mm}
  B2-VGG19~[\textcolor{green}{14}]~\hspace{15mm}
  HRN~[\textcolor{green}{14}]~\hspace{23mm}
  Ours
  \end{center}
  \vspace{-2mm}
  \caption{Confusion matrices of GAR by nearest neighbor retrieval on the VolleyBall Dataset (VBD) in GAFL-PAF.
  Each row and column show the ground-truth of the GA label and the recognized GA, respectively.
  While the results obtained by ``B2-VGG19'' and ``Ours'' are only shown in Fig.~\textcolor{red}{4} of the main paper, this figure shows the results obtained by all methods used in a comparison.
  }
  \label{fig:cm_usl_all_supp}
\end{figure*}

\subsubsection{Group Activity Recognition}

In this section, we show additional results of group activity recognition using 1-nearest neighbor classification, which is shown in Sec.~\textcolor{red}{4.4.2} of the main paper.

\paragraph{Confusion matrices in GAFL-PAC.}
As with the results in GAFL-PAF shown in Fig.~\textcolor{red}{4} of the main paper, confusion matrices in GAFL-PAC on the volleyball and collective activity datasets are also shown in Fig.~\ref{fig:cm_sl_all}.
As shown in Fig.~\ref{fig:cm_sl_all} (top), our method is better in all group activity classes on the volleyball dataset except for L-set in Dual-AI.
In the L-set, however, the accuracy difference between ``Dual-AI'' and ``Ours'' is small (i.e., 0.01 \%).
Furthermore, ``Ours'' achieves high-performance gain in L-pass, R-pass, L-winpoint, and R-winpoint compared with Dual-AI.
The average performance gain in these four group activity classes is 0.46 \%, so the performance drop in L-set (i.e., 0.01 \%) can be regarded as relatively small.
This is because our method learns visually subtle but important differences between these group activities, as demonstrated in the main paper (e.g., Fig.~\textcolor{red}{6}).
  
Figure~\ref{fig:cm_sl_all} (bottom) shows that our method is the best in all group activity classes on the collective activity dataset.
In particular, while the previous methods get many false negatives in Waiting, our method correctly recognizes Waiting.
We can interpret this reason that some Waiting scenes are visually similar to specific Moving scenes in which the movement of people is relatively small.
Such visually subtle differences are well discriminated in our method, as mentioned in the main paper.

\paragraph{Confusion matrices in GAFL-PAF.}
In addition to the results obtained by ``B2-VGG19'' shown in Fig.~\textcolor{red}{4} of the main paper, we further show the results obtained by ``B1-Compact128'' and ``HRN'' in Fig.~\ref{fig:cm_usl_all} of this supplementary material.
Regarding the additional results, we also see the same superiority of our method (i.e., Spike activity is well recognized in VBD, and false negatives in Moving are quite low on CAD), as noted in the main paper.

\begin{figure}[t]
  \begin{center}
  \includegraphics[width=\columnwidth]{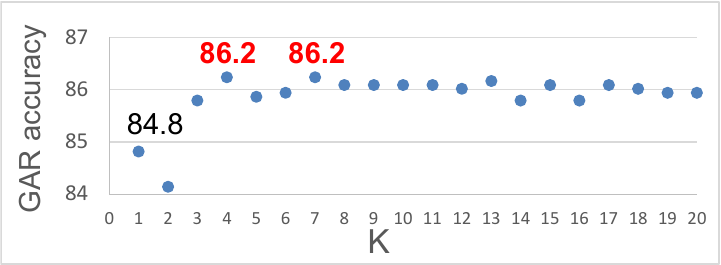}
  (a) VolleyBall Dataset (VBD)
  \includegraphics[width=\columnwidth]{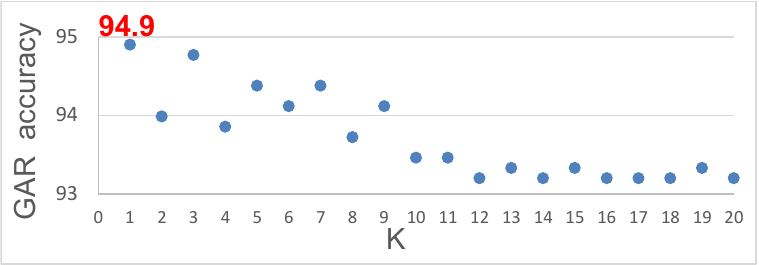}
  (b) Collective Activity Dataset (CAD)
  \end{center}
  \vspace{-4mm}
  \caption{GAR accuracy curve by the KNN classification in GAFL-PAC on the VolleyBall Dataset (VBD) and Collective Activity Dataset (CAD).
  $K$ changes from 1 to 20 in our experiment.
  }
 \label{fig:knn_sl_all}
\end{figure}

\paragraph{KNN for Group Activity Recognition in GAFL-PAC.}
In addition to the results of Group Activity Recognition (GAR) obtained by the 1-nearest neighbor classification shown in Figs.~\ref{fig:cm_sl_all} and~\ref{fig:cm_usl_all}, the results obtained with other neighbor numbers (e.g., 2) on the volleyball and collective activity datasets in GAFL-PAC are also shown in Fig.~\ref{fig:knn_sl_all}.
In our experiments, $K$ is changed from 1 to 20 for the $K$-nearest neighbor classification.

Fig.~\ref{fig:knn_sl_all} (a) shows that $K\geq3$ achieves better GAR accuracy than $K=1$ on the volleyball dataset. 
Specifically, the best results obtained by $K=4,7$ are 1.4$\%$ better than that of $K=1$. 
The results show that using KNN for GAR is simple but effective for accuracy. 
Results on the collective activity dataset shown in Fig.~\ref{fig:knn_sl_all} (b) show that the result obtained by $K=1$ (i.e., 94.9) is the best.
The results indicate that the 1-nearest neighbor classification shown in Fig.~\ref{fig:cm_sl_all} (bottom) is accurate enough in GAFL-PAC on the collective activity dataset.

\begin{figure}[t]
  \begin{center}
  \includegraphics[width=\columnwidth]{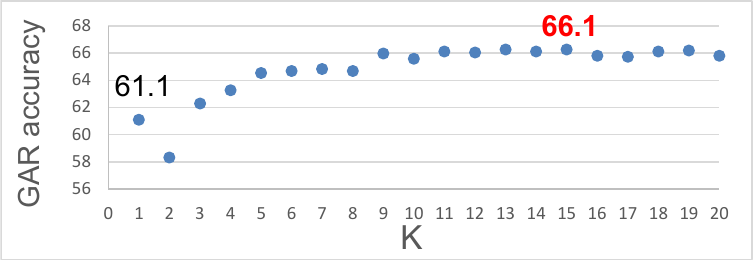}
  (a) VolleyBall Dataset (VBD)
  \includegraphics[width=\columnwidth]{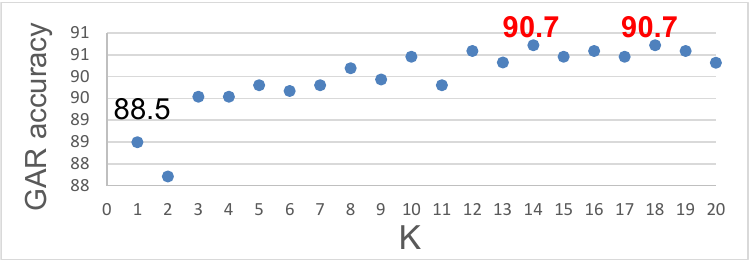}
  (b) Collective Activity Dataset (CAD)
  \end{center}
  \vspace{-4mm}
  \caption{GAR accuracy curve by the KNN classification in GAFL-PAF on the VolleyBall Dataset (VBD) and Collective Activity Dataset (CAD).
  $K$ changes from 1 to 20 in our experiment.
  }
 \label{fig:knn_usl_all}
\end{figure}

\paragraph{KNN for Group Activity Recognition in GAFL-PAF.}
As with the results above, the results obtained on the volleyball and collective activity datasets in GAFL-PAF are also shown in Fig.~\ref{fig:knn_usl_all}.
Figure~\ref{fig:knn_usl_all} shows that $K\geq3$ achieves better GAR accuracy than $K=1$ on the volleyball and collective activity datasets.
In particular, the best result obtained by $K=15$ on the volleyball dataset is 5.0$\%$ better than $K=1$.
The results indicate that the KNN is more effective when the performance obtained by $K=1$ is not highly accurate (i.e., 61.1 $\%$ on VBD in GAFL-PAF) compared with the accurate results (i.e., 84.8 $\%$ on VBD in GAFL-PAC).
This difference may come from the fact that a more abstract supervision signal (i.e., person appearance features) is used to learn in GAFL-PAF, so the GAF includes redundant information for representing manually annotated group activity classes.
For such GAF, ensembling ($K\geq2$) is more effective due to its robustness.

\begin{figure*}[t]
  \begin{center}
  \includegraphics[width=\textwidth]{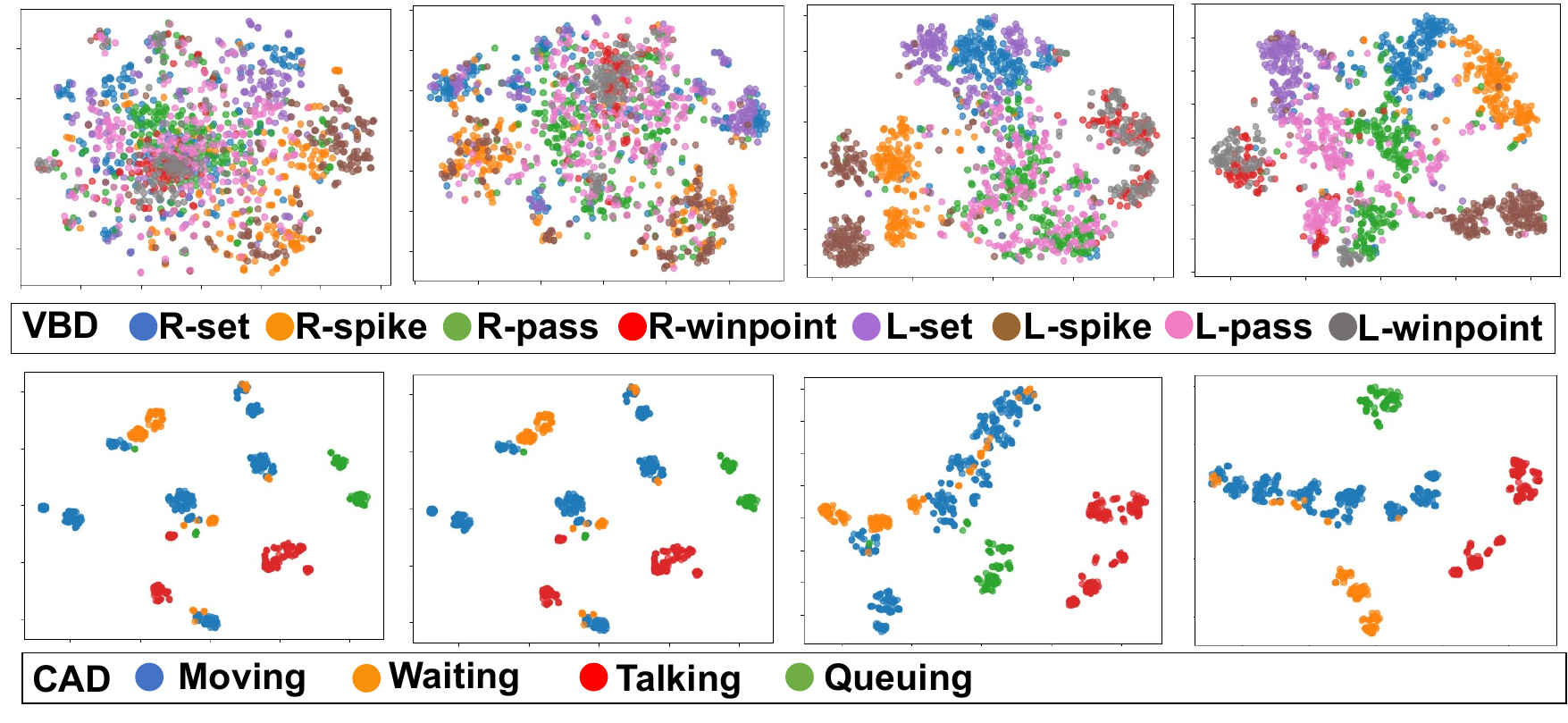}\\
  HiGCIN~[\textcolor{green}{37}]~\hspace{27mm}
  DIN~[\textcolor{green}{39}]~\hspace{27mm}
  Dual-AI~[\textcolor{green}{11}]~\hspace{27mm}
  Ours
  \end{center}
  \caption{
  Visualization of the learned GAF by t-SNE on the VolleyBall Dataset (VBD) and Collective Activity Dataset (CAD) in GAFL-PAC. 
  The color of each sample shows the annotated group activity class corresponding to each test sample. 
  Results obtained by ``Ours-grp'' are regarded as ``Ours'' in VBD and CAD.
  }
  \label{fig:latent_sl_all}
\end{figure*}

\subsubsection{Visualization of Learned Group Activity Feature}

\paragraph{GAF visualization in GAFL-PAC.}
As shown in Fig.~\textcolor{red}{5} of the main paper, the distribution of learned GAFs in GAFL-PAC is visualized in Fig.~\ref{fig:latent_sl_all}.
Figure~\ref{fig:latent_sl_all} shows that our method can learn the GAFs better than the other methods on the VolleyBall Dataset (VBD) and Collective Activity Dataset (CAD) in GAFL-PAC, as with the results in GAFL-PAF shown in the main paper.
The results on VBD shown in Fig.~\ref{fig:latent_sl_all} (upper) reveal that our GAF is useful for discriminating L-pass and R-pass, which are also mentioned in the above confusion matrices (Fig.~\ref{fig:cm_sl_all} in this supplementary material).
Figure~\ref{fig:latent_sl_all} (bottom) shows the results on CAD and indicates that our method learns GAF well in terms of representing visually similar Waiting and Moving activities.
The results are also validated in the above confusion metrics (Fig.~\ref{fig:cm_sl_all} in this supplementary material).

\begin{figure}[t]
  \begin{center}
  \includegraphics[width=\columnwidth]{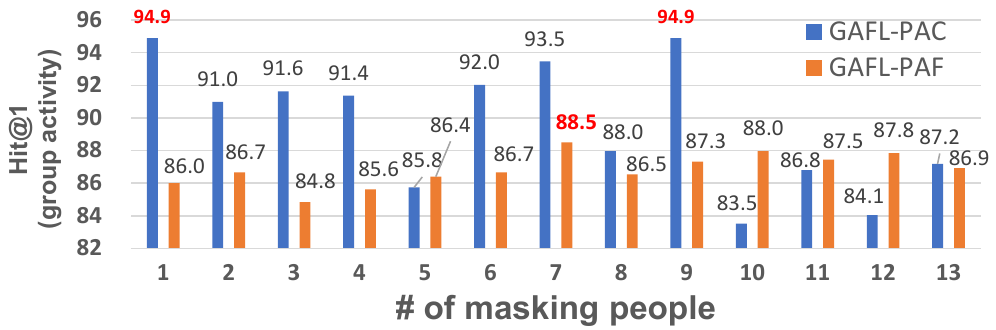}\\
  \end{center}
  \caption{Performance changes depending on the number of masking people on the Collective Activity Dataset (CAD).
  }
  \label{fig:mask_ratio_all_cad}
\end{figure}

\subsection{Detailed Analysis}
\label{subsection:det_analysis_supp}

\paragraph{Comparison of the number of masked persons.}
As shown in Fig.~\textcolor{red}{7} of the main paper, the performance changes depending on $N_{mask}$ on the Collective Activity Dataset (CAD) in GAFL-PAC and GAFL-PAF are shown in Fig.~\ref{fig:mask_ratio_all_cad}.
The results in GAFL-PAC (denoted by blue bars) show that our MPM is not important in the group activity retrieval performance.
%
This is because the GAF trained with person action labels is enough to represent group activities defined by the maximum number of person actions in CAD, even without our MPM.
In contrast, the results in GFL-PAF (denoted by orange bars) validate that our MPM improves the group activity retrieval performance.
Specifically, the results obtained by $\bm{N}_{mask}=7$ is 2.5$\%$ better than those without our MPM.
These results reveal that our MPM is also effective for learning GAF for such general scenes included in CAD.

\paragraph{Effect of location-guidance in our GAF learning.}

\begin{table}[t]
    \centering
    \caption{Effectivenss of our location-guidance in our GAF learning on the Collective Activity Dataset (CAD). 
    Results obtained by ``Ours-grp'' and ``Ours-ind'' are shown as ``Ours'' in GAFL-PAC and GAFL-PAF, respectively.
    }
    \begin{tabular}{l|l|c|c|c} \hline
    & \multicolumn{1}{c|}{\small{\shortstack{\\Retrieval type}}} & \multicolumn{1}{c|}{\footnotesize{\shortstack{\\Action set\\(IoU)}}} & \multicolumn{1}{c|}{\footnotesize{\shortstack{\\Action set\\(AF-IDF)}}} & \multicolumn{1}{c}{\footnotesize{\shortstack{\\Group\\activity}}}\\\cline{1-4}\hline
    & Method & Hit@1 & Hit@1 & Hit@1 \\ \hline\hline
    \multirow{2}{*}{\shortstack{GAFL-\\PAC}} & Ours w/o $\bm{F}_{loc}$ & 80.5 & 95.0 & 92.7 \\ \cline{2-5}
    & Ours & \red{81.8} & \red{96.1} & \red{94.9} \\ \hline\hline
    \multirow{2}{*}{\shortstack{GAFL-\\PAF}} & Ours w/o $\bm{F}_{loc}$ & 42.1 & 56.5 & 57.1 \\ \cline{2-5}
    & Ours & \red{67.6} & \red{83.7} & \red{88.5} \\ \cline{1-4}\hline
    \end{tabular}
    \label{table:det_abl_cad}
\end{table}

While the results of the ablation study for $\bm{F}_{loc}^{p}$ on VBD in GAFL-PAC and GAFL-PAF are shown in Table~\textcolor{red}{3} of the main paper, we further show the results on CAD in Table~\ref{table:det_abl_cad}.
On both GAFL-PAC and GAFL-PAF, ``Ours'' is better than ``Ours w/o $\bm{F}_{loc}^{p}$'' in all metrics.
%
%
%
In particular, the performance gain in the GAFL-PAF is larger than the one in GAFL-PAC.
We can interpret the reason as follows.
In GAFL-PAC, ``Ours w/o $\bm{F}_{loc}$'' learns person action distribution of a scene, as mentioned in Sec.~\textcolor{red}{4.5} of the main paper, is enough to represent group activities observed in CAD. 
This is because the location-related group activities (e.g., ``Waiting'' and ``Queuing'') can be represented by the distribution of person actions, which already includes location information in their class definition.
Furthermore, the group activities are defined by the maximum number of person actions in a scene on CAD.
Therefore, the group activities can be understood from person action distribution without $\bm{F}_{loc}$. 
In GAFL-PAF, however, person appearance feature distribution learned in ``Ours w/o $\bm{F}_{loc}$'' is not sufficient to represent group activities because people's appearances are similar to each other on CAD (e.g., the appearance of ``Waiting'' and ``Queuing'' actions are similar to each other).
To discriminate these similar appearances, our location guidance is effective.
For example, when people stand in line, their actions are likely to be regarded as ``Queuing'' even if their appearance features are similar to ``Waiting''.

In contrast to CAD, on VBD, our location guidance is crucial even in GAFL-PAC, as shown in Table~\textcolor{red}{3} of the main paper.
The results can be attributed to group activities where spatial relationships are meaningful (e.g., spiker and blocker are close in spike activity) on VBD.

\begin{figure}[t]
  \begin{center}
  \includegraphics[width=\columnwidth]{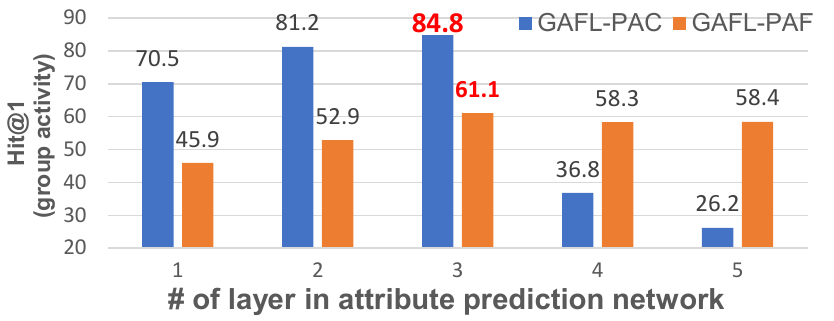}\\
  (a) VolleyBall Dataset (VBD)
  \includegraphics[width=\columnwidth]{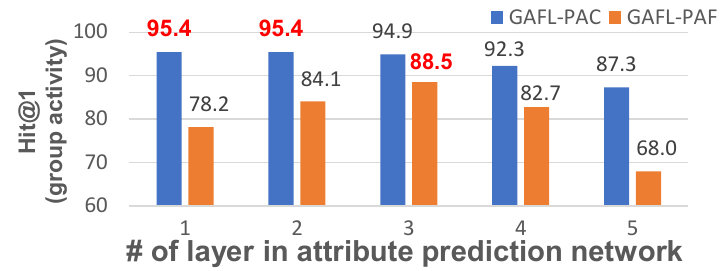}\\
  (b) Collective Activity Dataset (CAD)
  \end{center}
  \caption{Performance changes depending on the number of masking people on the VolleyBall Dataset (VBD) and Collective Activity Dataset (CAD).
  }
  \label{fig:act_layer_all}
\end{figure}

\begin{figure*}[t]
  \begin{center}
  \includegraphics[width=\textwidth]{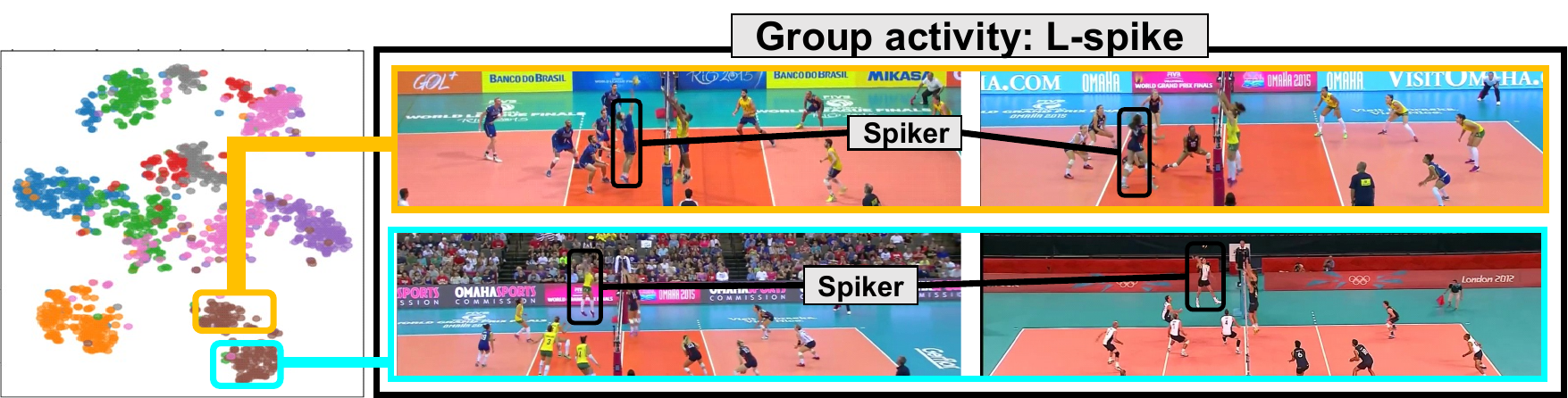}\\
  \end{center}
  \caption{Visualization of the learned GAF by t-SNE in GAFL-PAC. 
  The brown data points (i.e., ``L-spike'') are divided into two sub-categories based on the context (i.e., where the spiker is located in the left side court).
  }
  \label{fig:latent_sl_volley_l_spike}
\end{figure*}

\begin{figure*}[t]
  \begin{center}
  \includegraphics[width=\textwidth]{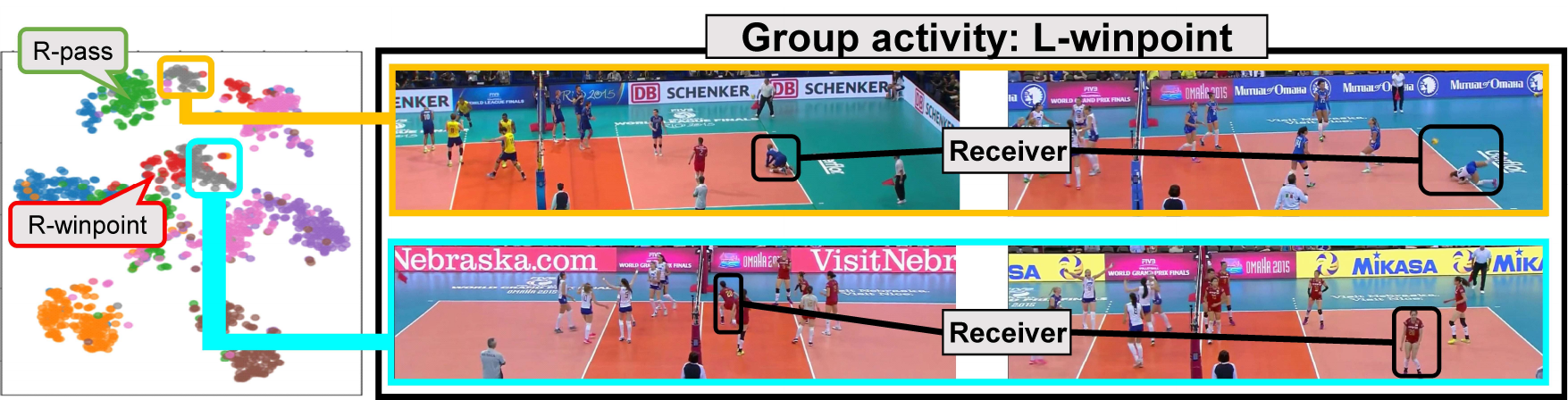}\\
  \end{center}
  \caption{Visualization of the learned GAF by t-SNE in GAFL-PAC. 
  The gray data points (i.e., ``L-winpoint'') are divided into two sub-categories based on the context (i.e., whether the receiver touched the ball or not).
  }
  \label{fig:latent_sl_volley_l_winpoint}
\end{figure*}

\begin{figure*}[t]
  \begin{center}
  \includegraphics[width=\textwidth]{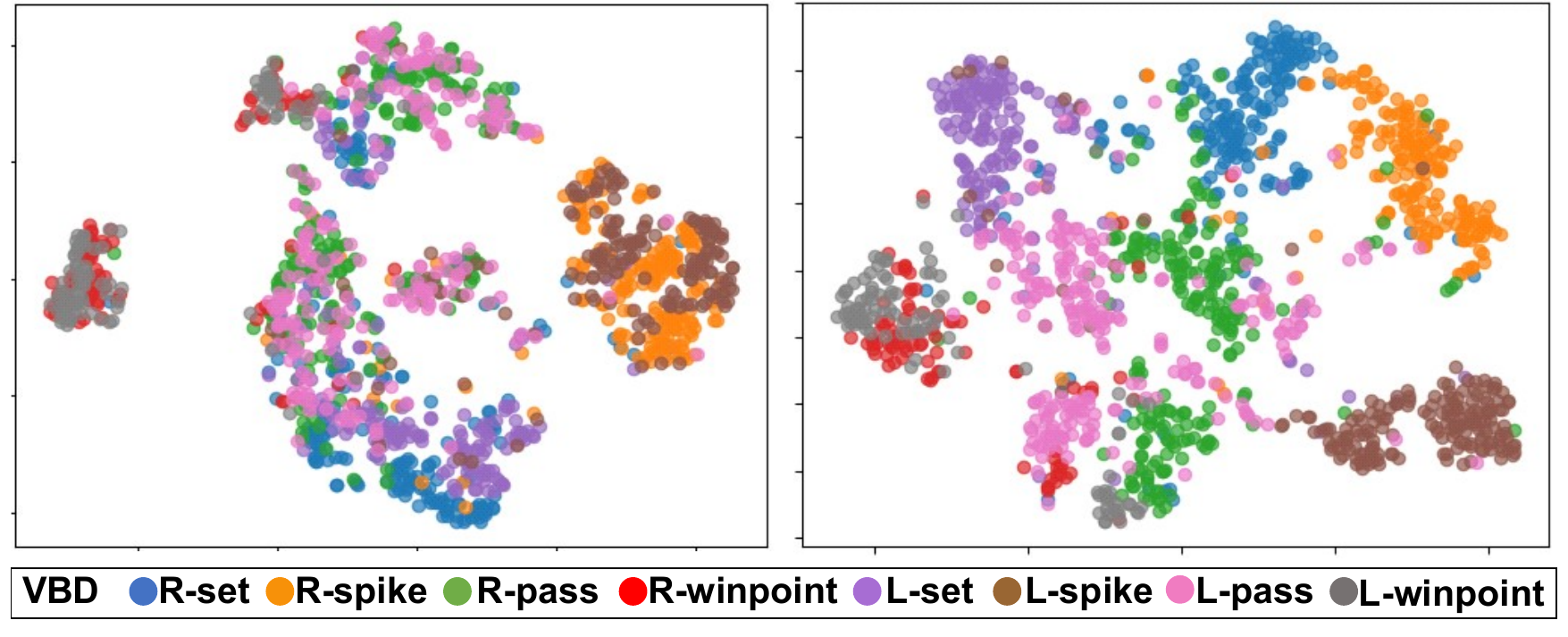}\\
  {\large
  Ours w/o $\bm{F}_{loc}$~\hspace*{70mm}
  Ours
  }
  \end{center}
  \caption{
  Effectiveness of our location-guidance in our GAF on the VolleyBall Dataset (VBD) in GAFL-PAC. 
  Results obtained by ``Ours-grp'' are regarded as ``Ours.''
  }
  \label{fig:latent_sl_volley_loc}
\end{figure*}

\paragraph{Optimal layer number for our attribute prediction network.}
Figure~\ref{fig:act_layer_all} shows that the retrieval performance changes depending on the number of layers in our attribute prediction network on the VolleyBall Dataset (VBD) and Collective Activity Dataset (CAD).
In general, such prediction performance is improved by increasing the number of layers.
However, we find that the performance decreases when the layer number is larger than four in Fig.~\ref{fig:act_layer_all}, while the performance increases until the layer number is three.
These results may come from the complexity of our person attribute prediction using the GAF compared with general person attribute prediction in which features of the target person are directly used.
The complexity may cause the model to overfit, so employing a shallow fully-connected network (e.g., 3-layer in this dataset) is adequate as our attribute prediction network.

The results on CAD in Fig.~\ref{fig:act_layer_all} (b) show that the performance is saturated even in the small number of layers (i.e., 1 and 2) in GAFL-PAC.
This is because the attribute prediction using GAF in GAFL-PAC on CAD is easier than the one on the VBD.
Therefore, such a shallow full connection network is enough for the attribute prediction network on CAD in GAFL-PAC.

\paragraph{Fine granularity of our GAF.}
As with the Fig.~\textcolor{red}{6} of the main paper, we further show additional examples that validate the fine granularity of our GAF in Figs.~\ref{fig:latent_sl_volley_l_spike} and~\ref{fig:latent_sl_volley_l_winpoint}.

In Fig.~\ref{fig:latent_sl_volley_l_spike}, we can see that the brown data points (i.e., L-spike) are divided into two sub-categories.
The two sub-categories differ in where the spiker hits the ball on the court.
Figure~\ref{fig:latent_sl_volley_l_winpoint} shows that the gray data points (i.e., L-winpoint) are split into two sub-categories due to whether the receiver touched the ball or not.
We further find the upper and bottom sub-categories are close to green (i.e., R-pass) and red (i.e., R-winpoint) data points, respectively.
The reason for this closeness can be interpreted as follows.
In the samples of this upper sub-category, the receiver touched a ball, so these samples are regarded as being similar to R-pass in which someone always touches a ball. 
In the samples of this bottom sub-category, the receiver focuses on but never touches the ball.
Therefore, the team on the right side can get the score if the ball goes out of the court.
This relationship with the scoring possibility in the bottom sub-categories makes the closeness with R-winpoint.

Furthermore, the effectiveness of our location-guidance for the fine granularity of our GAF is validated in Fig.~\ref{fig:latent_sl_volley_loc}.
This visualization shows that similar but subtly different group activities (e.g., the location of spiker is different in R-spike and L-spike while both represent spike activity) are not separated in ``Ours w/o $\bm{F}_{loc}$.''
From the results, we can interpret that our location-guidance is essential for learning where the group activity is happening, as also confirmed in Fig.~\ref{fig:latent_sl_volley_l_spike}.

\end{document}